\documentclass{article}

\usepackage{microtype}
\usepackage{graphicx}
\usepackage{booktabs} 

\usepackage{multirow}
\usepackage{caption}
\usepackage{subcaption}
\usepackage{amsmath}
\usepackage{amssymb}
\usepackage{pifont}
\newcommand{\cmark}{\ding{51}}%
\newcommand{\xmark}{\ding{55}}%

\DeclareMathOperator{\domainx}{\mathcal{X}}

\newcommand{\Tau}{\mathcal{T}}

\usepackage{graphicx}

\usepackage{xcolor}

\usepackage{hyperref}

\usepackage[accepted]{icml2020}

\icmltitlerunning{Local Nonparametric Meta-Learning}

\begin{document}

\twocolumn[
\icmltitle{Local Nonparametric Meta-Learning}



\icmlsetsymbol{equal}{*}

\begin{icmlauthorlist}
\icmlauthor{Wonjoon Goo}{ut}
\icmlauthor{Scott Niekum}{ut}
\end{icmlauthorlist}

\icmlaffiliation{ut}{Department of Computer Science, University of Texas at Austin, Austin, Texas}

\icmlcorrespondingauthor{Wonjoon Goo}{wonjoon@cs.utexas.edu}

\icmlkeywords{Meta-Learning, Nonparametric, Regression, Machine Learning}

\vskip 0.3in
]



\printAffiliationsAndNotice{}  

\begin{abstract}
A central goal of meta-learning is to find a learning rule that enables fast adaptation across a set of tasks, by learning the appropriate inductive bias for that set.
Most meta-learning algorithms try to find a \textit{global} learning rule that encodes this inductive bias.
However, a global learning rule represented by a fixed-size representation is prone to meta-underfitting or -overfitting since the right representational power for a task set is difficult to choose a priori.
Even when chosen correctly, we show that global, fixed-size representations often fail when confronted with certain types of out-of-distribution tasks, even when the same inductive bias is appropriate.
To address these problems, we propose a novel nonparametric meta-learning algorithm that utilizes a meta-trained local learning rule, building on recent ideas in attention-based and functional gradient-based meta-learning.
In several meta-regression problems, we show improved meta-generalization results using our local, nonparametric approach, and achieve  state-of-the-art results in the robotics benchmark, Omnipush.
\end{abstract}

\section{Introduction}

Fast adaptation with a small number of examples is a key challenge in many machine learning applications.
Meta-learning \cite{schmidhuber1987meta-learning,thrun2012learning} or learning-to-learn has been proposed as a path toward this ability, often by learning an update rule that leverages useful inductive bias shared across a set of related tasks for fast adaptation.
It has been widely investigated for few-shot classification or regression problems \cite{Vinyals16MatchingNet,Ravi2017MetaLSTM,snell2017prototypical,finn2017MAML,garnelo18cnp,kim18anp,tossou2019adaDKL} as well as in the context of reinforcement learning \cite{duan2016rl2,Wang2016LearningTR,mishra2018SNAIL,clavera2018learning}.

Unfortunately, meta-learning often fails when the test-time tasks differ from those in the meta-training set, even in relatively minor ways that do not change the underlying structure of the task. For example, an agent might be meta-trained on small mazes and then encounter a larger maze (meta-scale shift); or it may learn about the physics of objects that move slowly and then encounter a faster-moving object (meta-range shift).
These failures often occur for two primary reasons: (1) the learned inductive bias in the update rule overfits to the meta-training tasks \cite{Jamal2019TAML} and (2) the update rule is typically represented with a predefined, fixed-size representation (e.g. initial parameters of a neural network \cite{finn2017MAML} or an encoder that converts a training set into a fixed-size task vector \cite{garnelo18cnp}), which cannot scale to a larger-scale task at test time.

While approaches like MAML \cite{finn2017MAML} are technically universal \cite{finn18UnivMAML}, the meta-training set can almost never be expected to perfectly cover the range of conditions that will be encountered at test time.
Therefore, it is critical to be able to train a robust update rule that can still function well on out-of-distribution tasks as long as the underlying inductive bias of the tasks does not change; in other words, it must have meta-generalization ability.

To this end, we propose a novel algorithm, Meta-Regression using Local updates for Out-of-distribution Tasks (MeRLOT), in order to better generalize to two specific classes of out-of-distribution tasks: meta-scale shift and meta-range shift.
Meta-range shift occurs when a latent variable in a task, such as amplitude or period in a sinusoidal regression problem, takes on values outside of the range observed during meta-training. 
Meta-scale shift refers to tasks whose key latent variables require more bits to represent than those that were observed in the meta-training set, but retain the same semantics and appropriate inductive bias---for example, mazes that are larger than those seen at meta-training time.

MeRLOT is designed to achieve better meta-generalization by training a \textit{local} adaptation rule (as opposed to a global update that modifies the entire function at once), which is applied independently to local functions fit around each labeled data point. This results in a set of local functions that are specialized to their immediate context and which can be combined via an attention mechanism to represent the full function.
Locality brings several advantages with respect to meta-generalization ability:
\begin{itemize}
\item Meta-scale shift can be handled naturally, due to the nonparametric nature of the approach. Since we regress a separate local function around every data point, the representational power is not predefined and scales with more data / larger tasks.  
\item Meta-range shift can also be managed.
Since the domain-appropriate inductive bias is now learned as part of a local (rather than global) learning rule, the regressed functions are grounded more to local data points, rather than a global prior. 
\item Discontinuous functions can be regressed more accurately with a local representation.
While regressing a discontinuous function with a single neural network can be very challenging, our model can naturally handle the problem by having a set of local functions regressing each part of a piecewise-continuous function separately.
\end{itemize}

MeRLOT performs local adaptation via functional gradient descent \cite{manson2000functional,xu2019metafun}, in which the functional gradient is calculated with an attention mechanism \cite{vaswani17Attention,kim18anp}.
Our approach combines the strengths of several recent methods, namely the iterative functional gradient approach of MetaFun \cite{xu2019metafun} and context-based self-attention of Attentive Neural Process (ANP) \cite{kim18anp}, in addition to introducing the notion of locality in the update rule.

We evaluate the performance of MeRLOT against several other recent meta-learning approaches on a set of carefully designed meta-regression problems that cover both meta-range shift and meta-scale shift: (1) a 1D regression problem that includes discontinuities, (2) a forward dynamics modeling problem in a continuous maze, and (3) a popular real-world robotics benchmark, Omnipush \cite{bauza19omnipush}. 
Across all experimental domains, we were able to confirm the superior adaptability of MeRLOT to out-of-distribution tasks, compared to prior meta-learning approaches.

\section{Related Work}

Meta-learning can be broadly categorized into three categories: optimization-based \cite{Ravi2017MetaLSTM,finn2017MAML}, metric-based \cite{Vinyals16MatchingNet,snell2017prototypical}, and model-based \cite{santoro16MANN,mishra2018SNAIL,kim18anp}.
In optimization-based methods, such as MAML \cite{finn2017MAML}, the inductive bias is embedded in the form of initial parameters, which can be finetuned with a few gradient steps and a small dataset.
By contrast, metric-based meta-learning learns a similarity metric or kernel that is shareable across tasks so that the classification is done in the nearest-neighbor fashion.
Model-based, or encoder-decoder based meta-learning algorithms interpret the meta-learning problem as a sequence-to-sequence problem; a small training dataset is encoded into a fixed-size task encoding with a permutation invariant encoder \cite{garnelo18cnp}, and the decoder conditioned on the task encoding solves the task.
While both optimization-based and model-based meta-learning can theoretically represent any update rule \cite{finn2017MAML}, it is unclear how to apply metric-based meta-learning, such as \citet{snell2017prototypical}, for meta-learning problems other than classification tasks. That said, metric-based algorithms have a significant advantage, in that they handle meta-scale shift naturally due to their non-parametric nature, while optimization- and model-based methods do not.
In this work, we aim to combine the strengths of all of these disparate approaches into a single algorithm.

Achieving better meta-generalization ability has been a recent topic of interest in the meta-learning community.
It has been addressed in various ways, such as limiting the number of adaptable parameters \cite{lee2018MTNet,zintgraf2018CAVIA}, or modeling a posterior distribution of parameters instead of a point estimate \cite{taesup18BayesianMAML,na2019ood,grant2018recasting,finn18probMAML}.
By contrast, we aim to achieve meta-generalization via non-parametricity and iterative local adaptation.
Additionally, our work can be combined with other recently proposed regularization methods \cite{Jamal2019TAML,guiroy2019understanding, yin2020metaMemorization}, which use additional regularization terms to prevent meta-overfitting.



Attention mechanisms have led to significant performance gains in applications ranging from natural language processing \cite{Bahdanau2015NTM,vaswani17Attention} to meta-learning \cite{Vinyals16MatchingNet,santoro16MANN}. They have been especially effective in few-shot classification tasks, since attention can find and retrieve the closely related items.
Attention is also used to enhance task encodings \cite{mishra2018SNAIL,kim18anp} in model-based meta-learning algorithms, by supporting query-conditioned task encodings.


\section{Preliminaries}

\subsection{Meta-Learning}

In meta-learning, we want to find a (parameterized) learning rule $\Phi$ that can quickly adapt to a new task $\Tau$ from a given task distribution $\Tau \sim p(\Tau)$. In supervised meta-learning, $\Tau$ consists of a context (training) set $C = \{c_i := (x_i,y_i)\}_1^m$, a test (query) set $T = \{(x_j,y_j)\}_1^n$, and a loss function $l$. The loss function of meta-learning is formally defined as:
\begin{equation}
    L(\Phi;p(\Tau)) = \sum_{(C,T,l) \sim p(\Tau)} \big[ \sum_{(x_j,y_j) \in T} l(\hat{y_j},y_j) \big],
    \label{loss}
\end{equation}
where $\hat{y_j} = \Phi(x_j;C)$. We then minimize the meta-objective to find an optimal learning rule $\Phi^*$ for the task distribution $p(\Tau)$.

\subsection{Attention}

When a set of key-value pairs $\{(k_i,v_i)\}$ and a query $q$ is provided, an attention module (layer) typically computes the relevance weight $w_i$ for each element with respect to the query, and the module outputs the weighted sum of values:
\begin{align}
    \text{Att}(K,V,q) = w_q V \text{ where }w_q = f_{\text{att}}(K,q), \label{eq:attn}
\end{align}
where $K$ and $V$ is a matrix containing keys and values as a row vector, and $w_q$ is a row vector consists of $w_i$s.

There are a few widely used attention functions $f_{\text{att}}$s, such as $f_{\text{Laplace}} = \text{softmax}([|k_1 - q|_1,\dots,|k_{|K|} - q|_1])$ or $f_{\text{DotProduct}} = \text{softmax}([\langle k_i,q \rangle,\dots,\langle k_{|K|},q \rangle])$ where $|K|$ is the number of rows in the matrix $K$.
Multi-headed attention is also commonly used, which applies dot product attention multiple times with different linear transformation matrices for keys $K$, values $V$, and a query $q$ and then merges the outputs with another linear transformation matrix \cite{vaswani17Attention}. 

It is common to stack multiple attention modules with  non-linear layers in between, for example, as in an encoder-decoder architecture \cite{vaswani17Attention}.
In this architecture, the encoder gets the same inputs for keys, values, and queries (self-attention) while the decoder gets the output of the encoder as keys and values with extra queries (cross-attention).
The $l$th layer of the encoder and the decoder can be formally written as:
\begin{align}
    &E^{(0)} = K, D^{(0)} = [q_0, \dots, q_{|D|}] \\
    &E^{(l)} = \text{NL}\Big(E^{(l-1)},\big[\text{Att}(E^{(l-1)},E^{(l-1)},e^{(l-1)}_i)\big]_{i=1}^{|E|} \Big) \label{full_tr_att_1}\\
    &D^{(l)} = \text{NL}\Big(D^{(l-1)},\big[\text{Att}(E^{(l)},E^{(l)},d^{(l-1)}_i)\big]_{i=1}^{|D|}\Big)\\
    &\text{NL}(X,\Delta X) = \text{Norm}\Big(X + \text{FFN}\big(\text{Norm}(X+\Delta X)\big)\Big) \label{full_tr_att_2}
\end{align}
where $\text{Norm}$ is a layer-normalization layer, $\text{FFN}$ is a pointwise feed-forward network, $e_i$ and $d_i$ is the $i$th row vector of $E$ and $D$, and $[\cdot]_{i=1}^N$ is the matrix building operation of stacking row vectors.


\subsection{Encoder-decoder based Meta-learning}

In encoder-decoder based meta-learning algorithms \cite{garnelo18cnp,kim18anp,xu2019metafun}, an encoder summarizes a context set $C$ to a task encoding $h$, and a decoder $\Phi$ generates an adapted function $f$ using $h$, i.e. $f = \Phi(h)$.

In the Conditional Neural Process (CNP) \cite{garnelo18cnp}, the task encoding is a fixed-size vector, leading to limited adaptation ability. The Attentive Neural Process (ANP) \cite{kim18anp} resolves this issue by having an infinite dimensional task encoding. It is done by having a query-dependent task encoding $r_x$, in which an entire set of $r_x$ becomes the encoding, i.e. $h = \{r_x\}_{x \in \domainx}$. In ANP, $r_x$ is computed with \textit{context-based} self-attention layers: $r_x = \text{Att}(E^{(l)},E^{(l)},x)$ where $E^{(0)} = \{c_i:=(x_i,y_i)\}_1^m$.

MetaFun is similar to ANP in that a function is represented with a set of $r_x$, each of which is used to describe a function at $x$ with a help of decoder $\Phi$: $f(x) = \Phi(r_x)$.
The key difference is that MetaFun computes $r_x$ \textit{iteratively} with functional gradient descent \cite{manson2000functional}, treating each $r_x$ as a functional representation.
Each functional gradient descent step consists of three smaller steps: a gradient $u_i$ at each context data point is calculated, and the gradients are smoothed with a kernel (or similarity metric) $k(\cdot,\cdot)$. Then, the smoothed updates are applied to each $r_x$ with a learning rate $\alpha$ to adapt a function:
\begin{align}
    u_i(t) &= u(x_i,y_i,r_{x_i}(t)) \label{update_rule_1} \\
    \Delta r_{x}(t) &= \sum_{(x_i,\cdot) \in C} k(x_i,x) u_i(t) \\
    r_{x}(t+1) &= r_{x}(t) - \alpha \Delta r_{x}(t) \label{update_rule_2}
\end{align}
where $r_{x}(0)$ can be constant-initialized or trained with another parameterized function.
Note that they calculate an update $u_i$ with a parameterized neural network, not a formal derivative.
All learnable components, $u, k, \Phi$ are jointly optimized with Equation~\ref{loss}.

\begin{figure*}[t]
    \centering
    \includegraphics[width=0.95\textwidth]{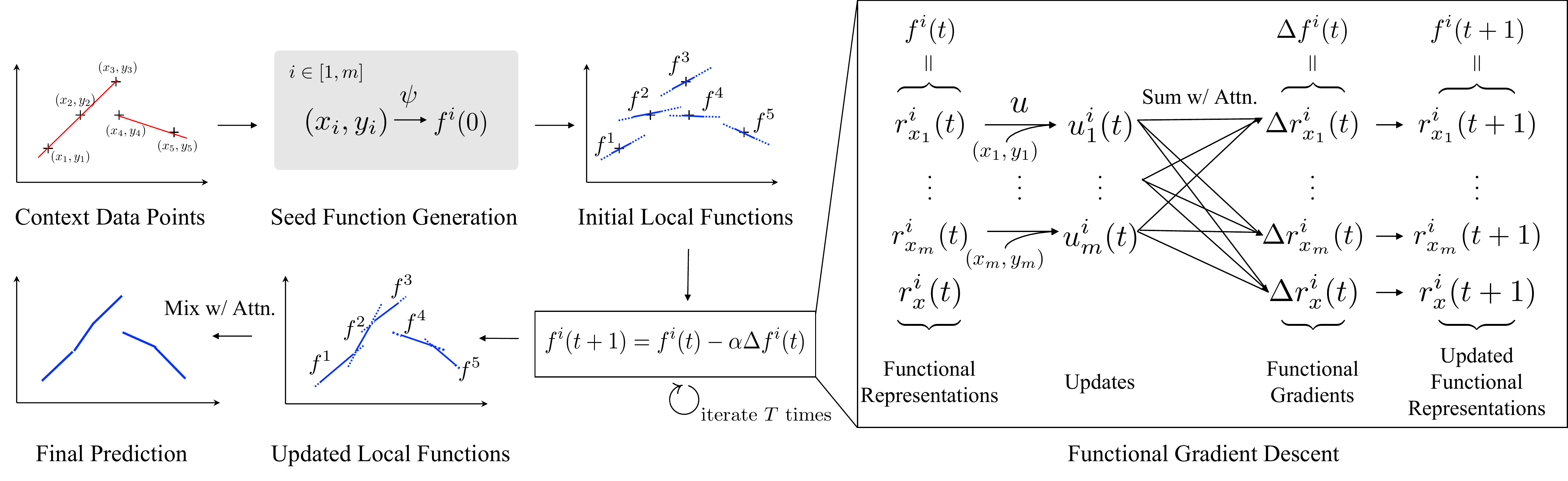}
    \caption{The model architecture of MeRLOT}
    \label{fig:model_architecture}
\end{figure*}

\section{MeRLOT}


We propose an algorithm: meta-regression using local updates for out-of-distribution tasks (MeRLOT). 
MeRLOT fits local functions around each data point 
via iterative functional gradient descent steps that utilize a parameterized update rule and context-based attention modules.
Figure~\ref{fig:model_architecture} illustrates the overall architecture\footnote{Code available \url{https://github.com/hiwonjoon/MeRLOT}.}.

The key characteristic of MeRLOT is that a set of functions $\{f^i\}_1^m$ is generated locally around every context data point $\{c_i:=(x_i,y_i)\}_1^m$, rather than fitting a single function $f$ to all data points. This process begins with a seed function generator $\psi$, which generates an initial function $f^i$ for each $c_i$. Since we represent a function as a set of functional representations $\{r_x\}_{x \in \domainx}$, $\psi$ generates a function that maps input $x$ to a functional representation $r_x^i(0)$ given context data point $c_i$:
\begin{equation}
    r_x^i(0) = \psi(c_i)(x).
\end{equation}
Then, we iteratively update each $r_x^i$ separately with the update rule given in Equation~\ref{update_rule_1}-\ref{update_rule_2}.
The final functional representations $r_x^i(T)$ for each query point is then interpreted with a decoder, and we get a set of adapted local functions $\{ f^i(x) := \Phi(r^i_x(T)) \}_1^m$.
To make a final prediction, we combine each local functions with a categorical distribution where the probability of $i$th function being selected is given by an attention value $k(x,x_i)$. When each function $f^i$ predicts a Gaussian distribution by predicting $\mu_i$ and $\sigma_i$ instead of point estimate to generate an uncertainty, the final prediction becomes a Gaussian mixture model:
\begin{gather}
    p(y=f^i(x)|x, \{c_i\}; \psi, u, k, \Phi) = k(x,x_i) \text{ or} \\
    p(y|x, \{c_i\}; \psi, u, k, \Phi) = \sum_i k(x,x_i) \mathcal{N}(\mu_i,\sigma_i).
\end{gather}

On top of iterative functional updates of local functions, MeRLOT also uses a context-dependent similarity metric $k$, similar to ANP.
Specifically, MeRLOT uses a full Transformer-style \cite{vaswani17Attention} attention module to build a context-dependent similarity metric $k$;
we first compute embedding vectors $d^{(L)}$s for a query point $x$ with the $L$-stack of encoder-decoder architecture introduced in Equation~\ref{full_tr_att_1}-\ref{full_tr_att_2}. We then use a dot-product attention to calculate an attention value between context data points $\{c_i:=(x_i,y_i)\}_{i=1}^{m}$ and the query $x$:
\begin{align}
    [k(x,x_1), ..., k(x,x_m)] = f_{\text{DotProduct}}(D^{(L)}_{1:m}, d^{(L)}_{m+1})
\end{align}
where $E^{(0)} = [c_1, \dots, c_m]$ and $D^{(0)} = [x_1,\dots,x_m,x]$.

Our approach integrates beneficial elements of both ANP (context-based attention) and MetaFun (iterative functional gradient updates), while additionally introducing locality. Together, these ingredients allow MeRLOT to better generalize to out-of-distribution tasks than previous meta-learning methods, including those we build upon.
The primary advantage of locality is that the prediction made by $f^i$ is more grounded to a context label $y_i$, in the sense that the value $y$ has shorter path to output in computation than in ANP.
Furthermore, since the update rule is local, it relies more on inductive biases that are local in nature, rather than global function priors that can fail to generalize under task shift.
There are four trainable components in MeRLOT: the seed functional generator $\psi$, the updater $u$, the decoder $\Phi$, and the context dependent similarity metric $k$. We use a multi-layer perceptron (MLP) for each, except for the similarity metric $k$ in which we use the encoder-decoder attention network.
Also, for the modules $\psi$ and $\Phi$, which generate a function instead of a vector, we cascade two MLPs to implement the module; the first MLP generates an embedding, which is sent to the next MLP with an extra input, i.e. $r^i_x(0) = \text{MLP}\big(x,\text{MLP}_{\psi}(x_i,y_i)\big)$.
For MLPs with multiple inputs, we concatenate the inputs.
When $f^i$ predicts a Gaussian distribution, the entire modules are optimized with the loss function shown in Equation~\ref{loss} with negative log-likelihood (NLL) as $l$. When $f^i$ predicts a point estimate, we train the modules with $l_2$ loss. We use $T=3$ and $\alpha=0.01$. 

\section{Experiments}


\subsection{1D Function Regression}

\begin{figure*}[t]
    \centering
    \begin{subfigure}{0.245\textwidth}
        \includegraphics[width=\textwidth]{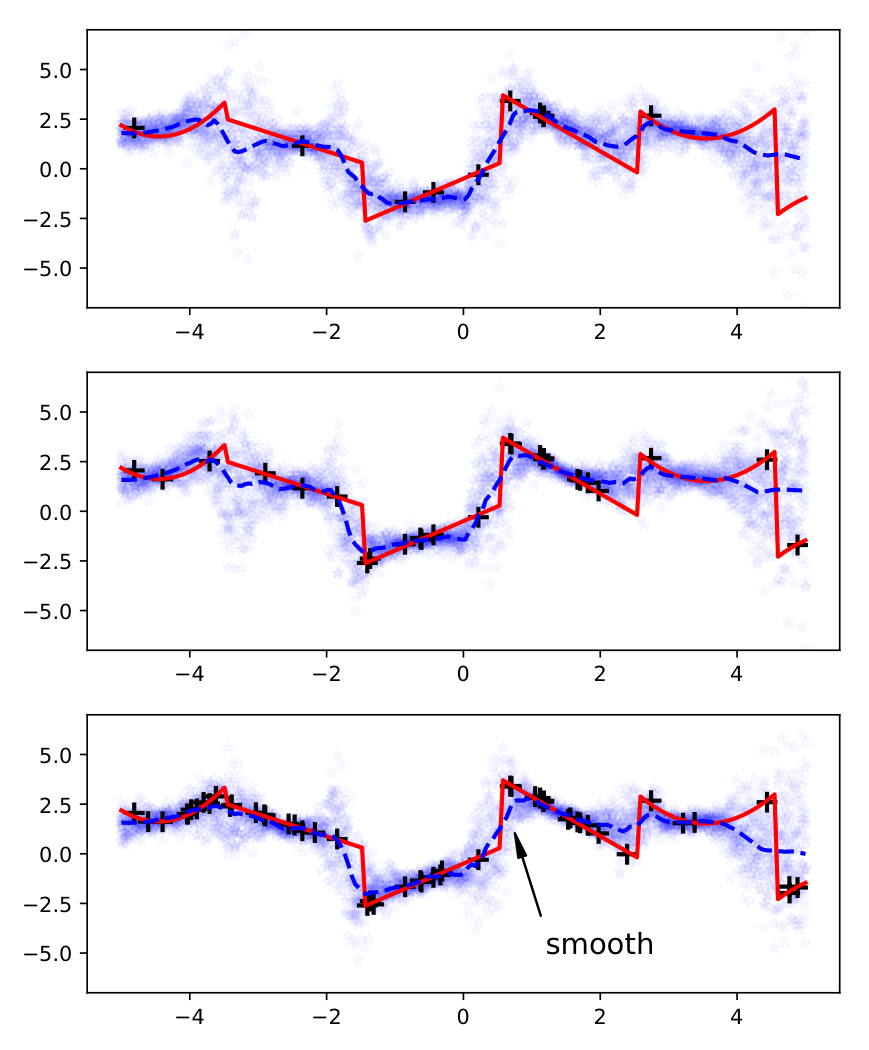}
        \caption{MAML}
        \label{fig:one_d_qual_maml}
    \end{subfigure}
    \begin{subfigure}{0.245\textwidth}
        \includegraphics[width=\textwidth]{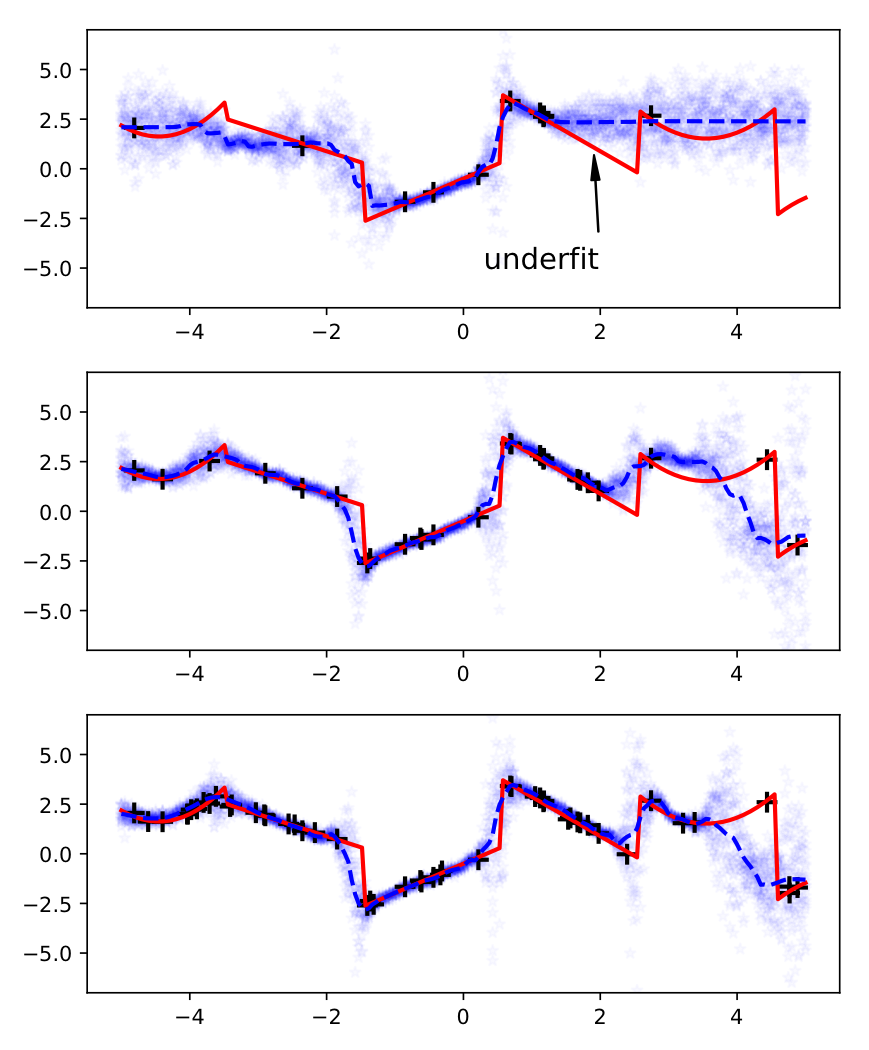}
        \caption{ANP}
        \label{fig:one_d_qual_anp}
    \end{subfigure}
    \begin{subfigure}{0.245\textwidth}
        \includegraphics[width=\textwidth]{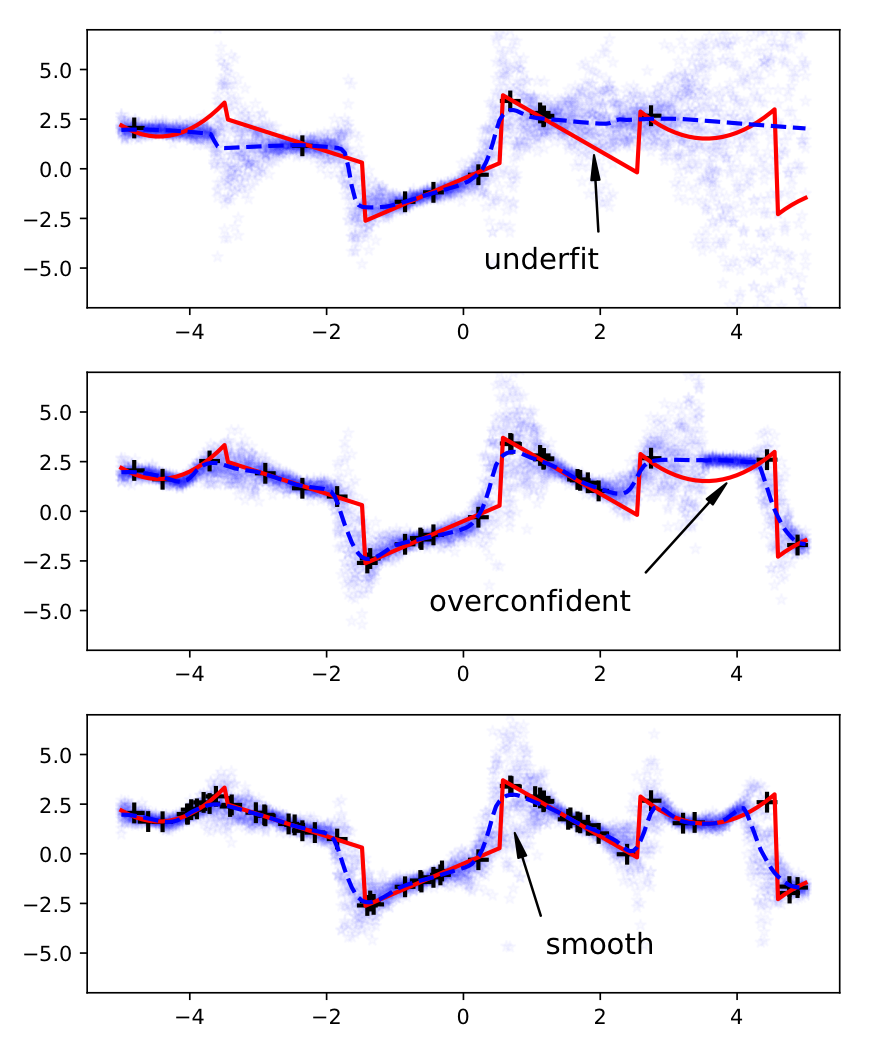}
        \caption{MetaFun}
        \label{fig:one_d_qual_metafun}
    \end{subfigure}
    \begin{subfigure}{0.245\textwidth}
        \includegraphics[width=\textwidth]{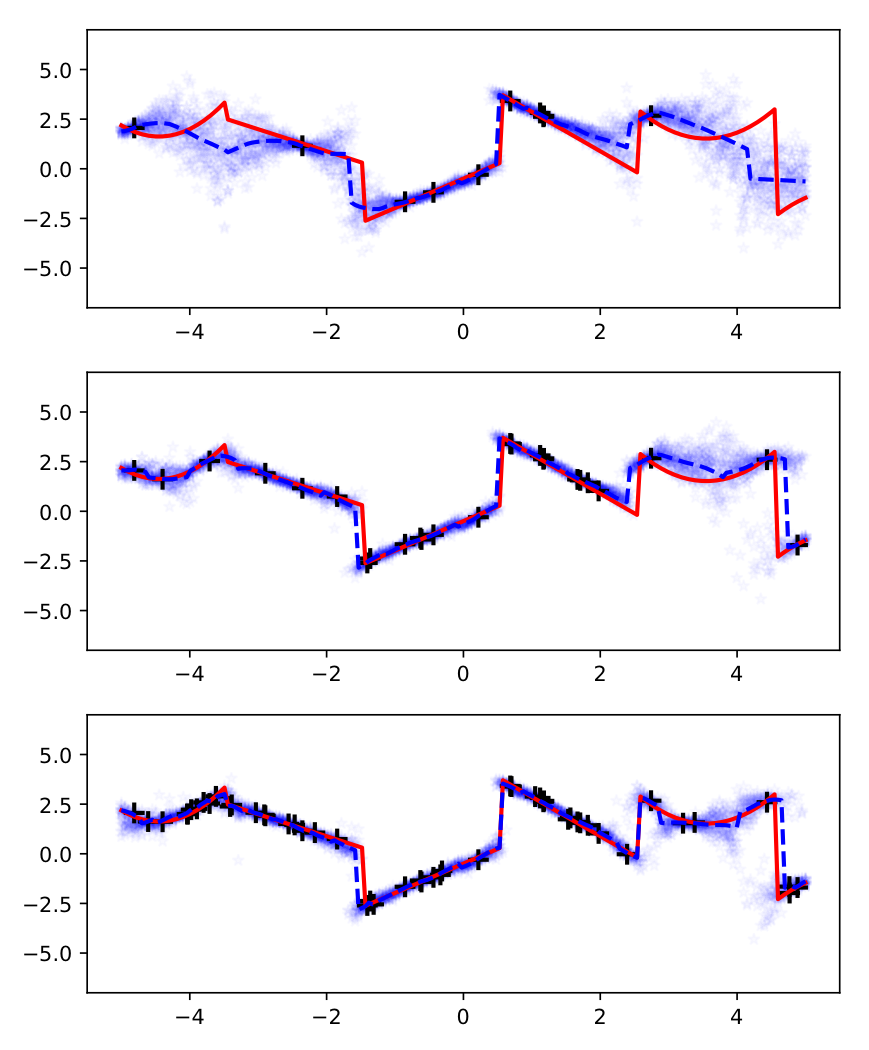}
        \caption{MeRLOT}
        \label{fig:one_d_qual_ours}
    \end{subfigure}
    \caption{1D function regression results for a meta-scale shift task with different number of context data points (from the top, 10, 25, and 50 data points are given). The ground-truth function is drawn as a solid red line, and the given context data points are drawn with a black plus symbol $+$. The mode of the predicted distribution is drawn as a dashed blue line. To represent the uncertainty of the prediction, 20 samples are drawn from the distribution for each x and those samples are drawn in blue dots. 
    More results included in the appendix.}
    \label{fig:one_d_qual}
\end{figure*}

We first examine meta-generalization in a 1D function regression problem that allows us to observe different behaviors of each meta-learning method visually.
We carefully design this problem to have the two types of out-of-distribution tasks---meta-range shift and meta-scale shift---and also consider target functions which include discontinuities.

The target functions are generated as follows:
we construct the union of 5 or 6 consecutive, non-overlapping intervals of size 2 (e.g. $[-6,-4], \cdots, [4,6]$) that contains the domain $[-5,5]$.
Then for each interval $[c-1,c+1]$, we define a linear or quadratic function of the form $y=a(x-c)+b$ or $y=a(x-c)^2+b$, where the coefficients are uniformly sampled from $[-3,3]$: $a, b \sim \text{unif}(-3,3)$.
Finally, these continuous pieces are concatenated to form the piecewise-continuous target function.
Therefore, to succeed in this meta-learning problem, a meta-trained learning rule must use the context data points to infer the boundaries and shape of each continuous piece of the function.
Figure~\ref{fig:one_d_qual} shows an example of a target function.


\textbf{Meta-range shift} We categorize the random target functions into three classes based on their underlying coefficients, $a$ and $b$.
We split the original range of coefficients $[-3,3]$ into a meta-training set $[-2,-1] \cup [1,2]$, an interpolation meta-test set $[-1,1]$, and an extrapolation meta-test set $[-3,-2] \cup [2,3]$. 

During meta-training and meta-testing, 7 to 15 data points are sampled uniformly in a randomly-generated interval of size 4, and the points are used as the context set $C$. Another 5 to 10 data points are sampled in the same interval to form the query set $T$.
Since the chosen interval can include the domain of two or three underlying continuous functions, the meta-learner only needs to regress those parts. For the context set, we added Gaussian observation noise of zero mean and standard deviation of $0.1$.

\textbf{Meta-scale shift} In this scenario, we uniformly sample $x$ from the full domain $[-5,5]$ for the context set instead of an interval of size 4 as was done in the meta-range shift experiment.
Note that this increases the minimum description length of the target concept; the learner must regress to five or six continuous pieces instead of two to three.
We use target functions generated with the same constant range of meta-training. The performance on this test scenario is measured with 100 query points sampled uniformly from $[-5,5]$ while the number of given context data points changes from 5 to 100.

As evaluation metrics, we use root mean squared error (RMSE) to measure the accuracy of predictions, as well as negative log-likelihood (NLL), which is more useful for evaluating the accuracy of the predicted uncertainties. We reported the averaged RMSE and NLL over 1,000 tasks. For RMSE, we use a mean of 30 sample predictions as a final prediction.
We consider three recent meta-learning approaches: Model Agnostic Meta-Learning (MAML) \cite{finn2017MAML}, as well as ANP \cite{kim18anp} and MetaFun \cite{xu2019metafun}, which MeRLOT builds upon.
The hyperparameters for each network are provided in the appendix, and the code is also available.

\begin{figure}
    \centering
    \begin{subfigure}{0.23\textwidth}
        \includegraphics[width=\textwidth]{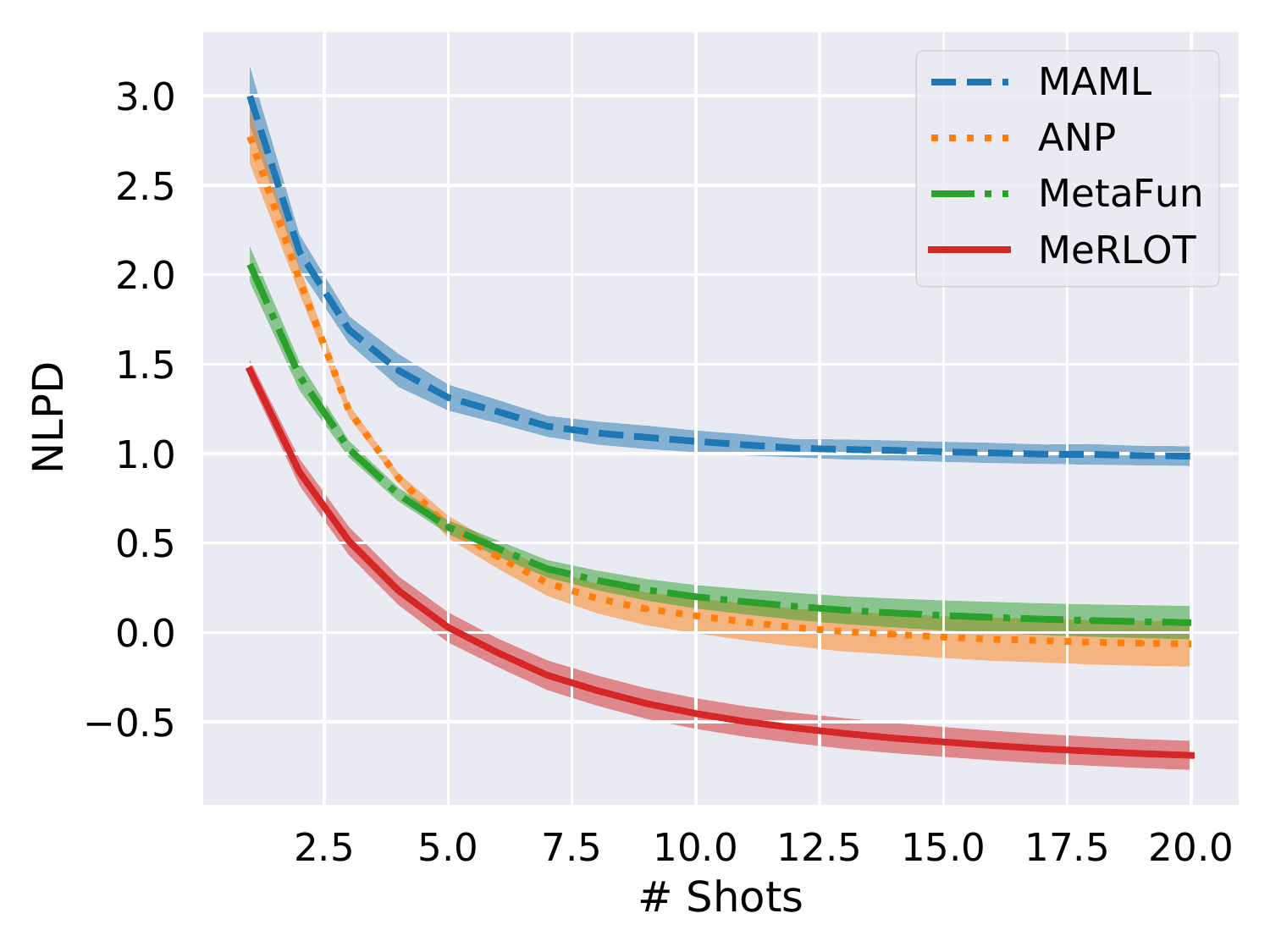}
        \caption{NLL}
    \end{subfigure}
    \begin{subfigure}{0.23\textwidth}
        \includegraphics[width=\textwidth]{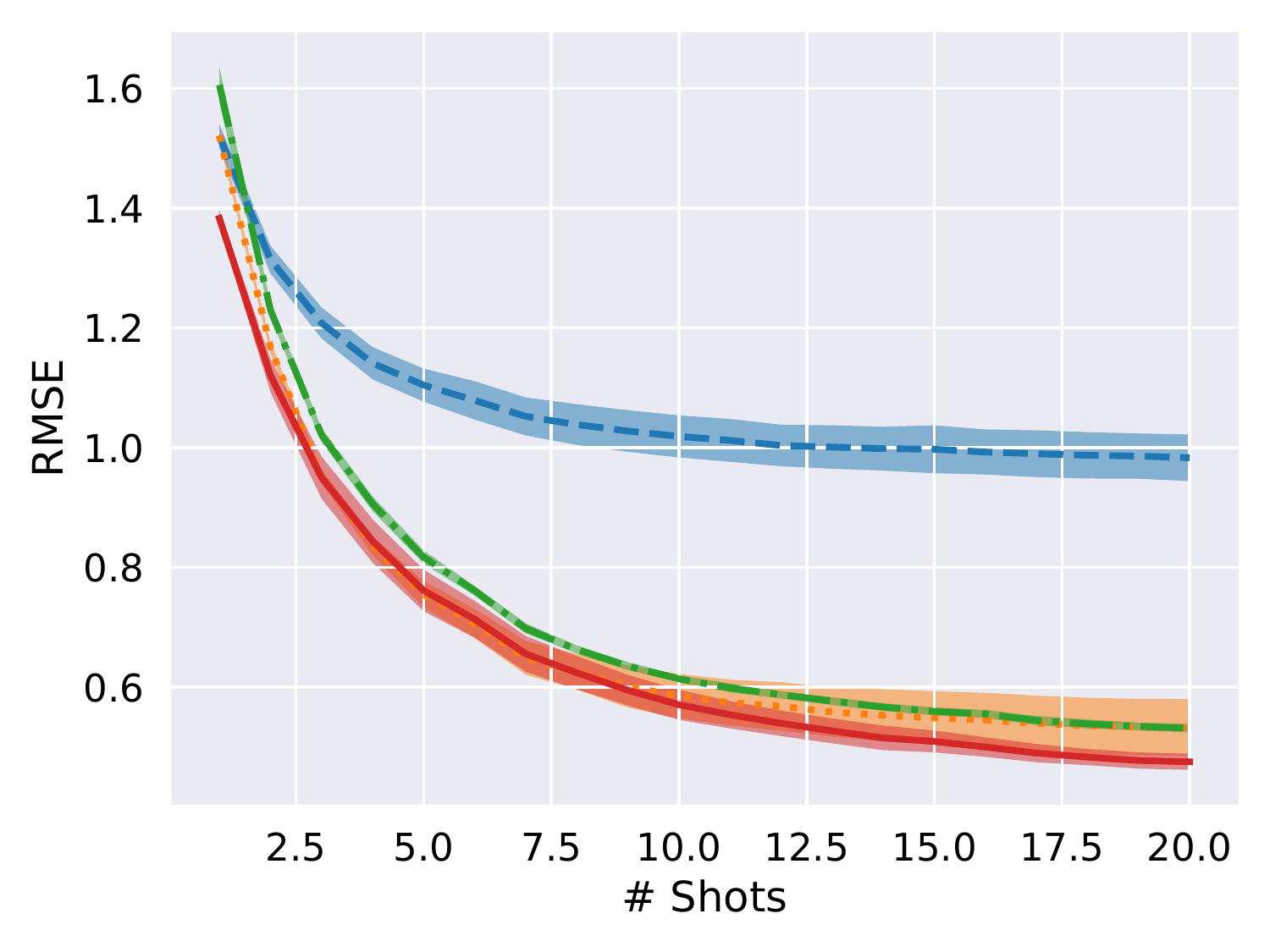}
        \caption{RMSE}
    \end{subfigure}
    \caption{Meta-scale shift results of 1D function regression. A shaded uncertainty band represents standard deviation of three different runs.
    }
    \label{fig:one_d_scale_change}
\end{figure}

\begin{table*}[t]
\centering
\caption{Quantitative results for two out-of-distribution meta-test scenarios for 1D function regression. The experiment is repeated three times with different random seeds, and we reported the average of the runs. The top rows of the table display results for baselines and MeRLOT; lower rows show ablations of MetaFun and MeRLOT with self attention added and seed generation removed, respectively. For ease of comparison, the left side of the table displays whether each method includes self attention, iterative updates, local functions, and seed generation ($\psi$).  Full results with standard deviations are included in the appendix.}
\vskip 0.1in
\begin{tabular}{@{}lccccrrrrrrrr@{}}
\toprule
 & \multirow{3}{*}{\begin{tabular}[c]{@{}c@{}}Self\\ Att.\end{tabular}} & \multirow{3}{*}{\begin{tabular}[c]{@{}c@{}}Iter.\\ Upd.\end{tabular}} & \multirow{3}{*}{\begin{tabular}[c]{@{}c@{}}Loc.\\ Fns.\end{tabular}} & \multirow{3}{*}{$\psi$} & \multicolumn{4}{l}{Meta-range shift} & \multicolumn{4}{l}{Meta-scale shift} \\
Method & & & & & \multicolumn{2}{l}{Interpolation} & \multicolumn{2}{l}{Extrapolation} & \multicolumn{2}{l}{10 context pts.} & \multicolumn{2}{l}{50 context pts.} \\ 
\multicolumn{1}{c}{} &  &  &  &  & NLL & RMSE & NLL & RMSE & NLL & RMSE & NLL & RMSE \\
\midrule
MAML & N/A & N/A & N/A & N/A &
0.828 & 0.528 & 1.806 & 1.686 & 2.127 & 1.314 & 1.068 & 1.019 \\
ANP & \cmark & \xmark & N/A & N/A & 
0.883 & 0.425 & 1.920 & 1.331 & 1.970 & 1.167 & 0.093 & 0.585 \\
MetaFun & \xmark & \cmark & \xmark & \xmark & 
0.524 & \textbf{0.316} & 1.563 & 1.297 & 1.433 & 1.230 & 0.199 & 0.614 \\
MeRLOT & \cmark & \cmark & \cmark & \cmark & 
\textbf{-0.059} & 0.331 & \textbf{1.104} & \textbf{1.143} & \textbf{0.896} & \textbf{1.121} & \textbf{-0.453} & \textbf{0.570} \\ 
\midrule
MetaFun w/ SA & \cmark & \cmark & \xmark & \xmark & 
0.272 & \textbf{0.310} & 1.195 & 1.211 & 1.716 & 1.315 & 0.918 & 0.796 \\
MeRLOT w/o $\psi$ & \cmark & \cmark & \cmark & \xmark & 
\textbf{-0.072} & \textbf{0.315} & \textbf{0.869} & 1.328 & 1.072 & 1.125 & 0.033 & 0.640 \\
\bottomrule
\end{tabular}
\label{tab:1dregression}
\vskip -0.1in
\end{table*}
The quantitative results are shown in Table~\ref{tab:1dregression} and Figure~\ref{fig:one_d_scale_change}.
MeRLOT surpassed all other approaches with a large margin in the NLL metric for all meta-testing scenarios, as well as the RMSE metric for all meta-testing scenarios except the interpolation task set.
Notably, we observe a better utilization of context data points in Figure~\ref{fig:one_d_scale_change}, in which RMSE and NLL continue to decrease with additional data more significantly for MeRLOT than other methods.
This corresponds to the qualitative results in Figure~\ref{fig:one_d_qual}; MeRLOT can regress an entire function more precisely, whether data is scarce or abundant, compared to other methods that underfit.
For instance, the underfitting behaviors of ANP and MetaFun (the first row in Figure~\ref{fig:one_d_qual}) demonstrates that both methods fail to learn the inductive bias of the task that a continuous piece has a domain length of 2.
We hypothesize the improved performance of MeRLOT results from (1) its ability to retrieve related data points from the given context better than MetaFun, due to locality and the context-based similarity metric and (2) the fact that it has a shorter computation path for prediction than ANP, due to the use of functional gradients and the seed function generator.
Also, the failure of MAML in meta-scale shift scenario corresponds to our conjecture on having fixed representation power; when the scale of a task increases, the representation power needs to increase as well, but the power of MAML is capped by the size of parameters of a neural network.

Moreover, we confirm the advantage of regressing local functions for handling discontinuities.
In Figure~\ref{fig:one_d_qual}, MeRLOT handles a discontinuity very precisely with sharp predictions while the predictions of MAML and MetaFun are overly smooth.
We attribute this to the fact that MAML uses a single neural network to model the function globally and MetaFun smooths the $u_i$ updates with a similarity metric that depends only on $x$. In contrast, MeRLOT models multiple local functions, and aggregates the updates adaptively based on a self-attentive similarity metric that considers both $x$ and $y$.

\textbf{Ablation Study}
We conducted the same experiments with ablated networks to understand the behavior of each component of MeRLOT. 
We tried two different ablations. First, we supplement the original MetaFun algorithm with a self-attention module of MeRLOT, which considers the full set of context data points in calculating similarity. Second, we disable the seed function generator $\psi$ of MeRLOT; instead of generating $f^i$ with the context data point $c_i = (x_i,y_i)$, a seed function $f^i$ is generated only with $x_i$. The results are shown in Table~\ref{tab:1dregression}.

\begin{figure*}[t]
    \centering
    \includegraphics[width=\textwidth]{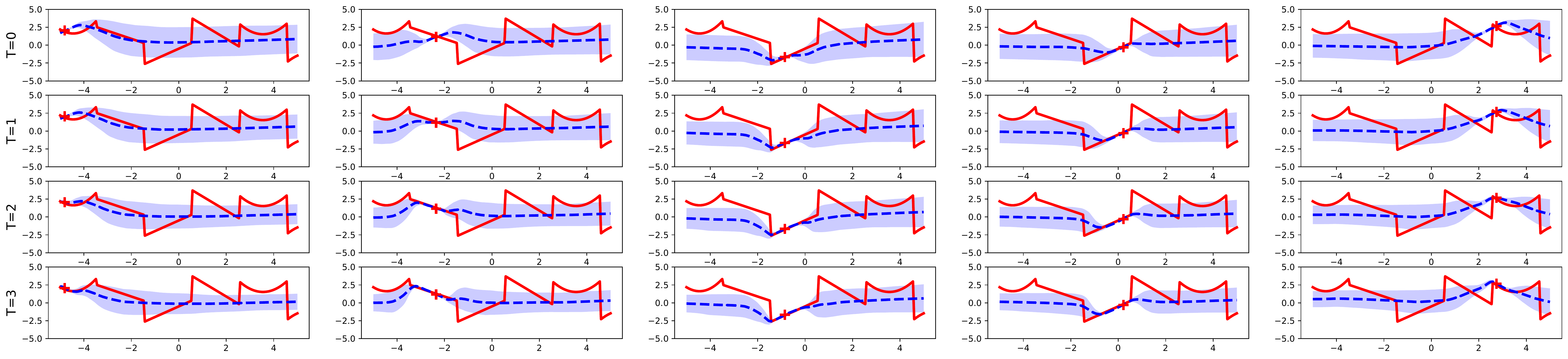}
    \caption{
    Visualization of local adaptation of MeRLOT.
    Each row represents different update iterations $t$, and each column represents the different local function $f^i(t)$ based on each context data point $(x_i,y_i)$. Each context point is drawn with a red plus symbol $+$.
    The mean of $f^i(t)$ is drawn as a blue line with a shaded uncertainty band representing standard deviation of the prediction.
    We used a meta-scale shift set with the context size of 50, and we display 5 local functions among 50 local functions.
    }
    \label{fig:locality}
    \vskip -0.1in
\end{figure*}

The performances of both ablated networks are better than standard MetaFun in meta-range shift tasks. This result indicates the advantage of using the self-attentive similarity metric and adapting local functions to deal with range shift.
However, the two ablated networks perform worse in meta-scale shift tasks than MeRLOT. This indicates the importance of the seed function generator $\psi$ in supporting meta-generalization to larger tasks. By generating good, locally accurate initial functional representations $r^i_x(0)$, the seed generator allows the updater to learn to perform small, \textit{local} updates accurately.
Corresponding to this qualitative result, the visualization of the iterative update process also verifies the local update property of MeRLOT;
in Figure~\ref{fig:locality}, we can observe that $\psi$ generates a proper initial function (top row, when T=0), which crosses a given context data point $(x,y)$, and each function is only locally adjusted to regress its dedicated surroundings.
However, when the initial local functions are not initialized with $\psi$ (see Figure~\ref{fig:locality:ablated}), the updater $u$ appears to alter each function globally, losing the advantage gained from regressing multiple local functions.


\subsection{Maze}

We next explore a forward dynamics prediction problem of a ball (point mass) in a 2D-maze environment. The environment is implemented using the Mujoco physics simulator, where the state space consists of $x,y$ positions and velocities, and the action space consists of forces in each $x,y$ direction.
While the movement of the ball follows simple Newtonian mechanics, the maze configuration is not directly observable. Hence, a learner has to infer the existence of walls from a context set of actions and resulting states to predict forward dynamics correctly.

For meta-training, we generate 1,000 configurations of mazes having size between $7\times7$ and $9\times9$. Then, for each maze configuration, 100 episodes (trajectories) of length 300 are generated with a random policy that applies force at random direction every step.
We sample a 50-length trajectory snippet from one of the trajectories, and we used the first 30 transition tuples for a context set and the rest for a test set.
Each transition tuple consists of state, action, next state, and we concatenate state and action to make input $x$ while output $y$ is defined as a difference between next state and a current state. Both $x$ and $y$ are normalized to have a standard deviation of 1 for each dimension. In this experiments, we directly predicted a target $y$ instead of predicting a distribution since the simulator is deterministic.

\begin{figure}[t]
    \centering
    \begin{subfigure}{0.3\textwidth}
        \centering
        \includegraphics[width=0.345\textwidth]{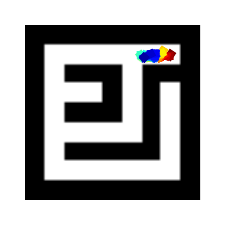}
        \includegraphics[width=0.345\textwidth]{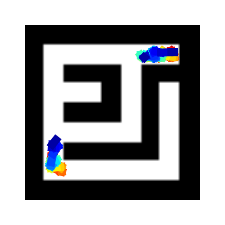}
        \caption{Given context trajectories}
    \end{subfigure}
    \begin{subfigure}{0.15\textwidth}
        \centering
        \includegraphics[width=0.69\textwidth]{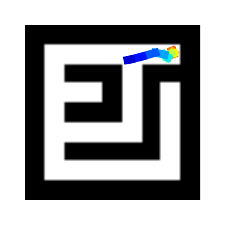}
        \caption{Ground-truth}
    \end{subfigure}
    \begin{subfigure}{0.155\textwidth}
        \includegraphics[width=0.48\textwidth]{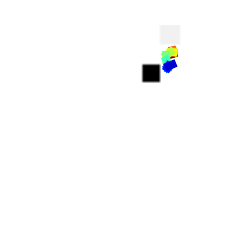}
        \includegraphics[width=0.48\textwidth]{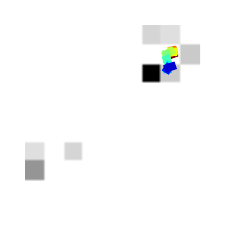}
        \caption{MAML}
    \end{subfigure}
    \begin{subfigure}{0.155\textwidth}
        \includegraphics[width=0.48\textwidth]{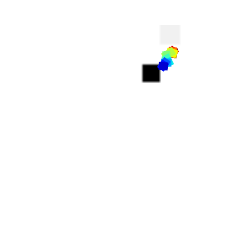}
        \includegraphics[width=0.48\textwidth]{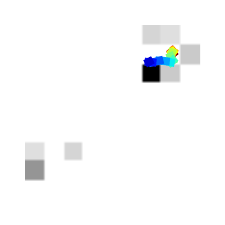}
        \caption{ANP}
    \end{subfigure}
    \begin{subfigure}{0.155\textwidth}
        \includegraphics[width=0.48\textwidth]{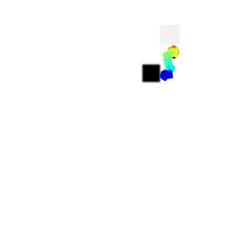}
        \includegraphics[width=0.48\textwidth]{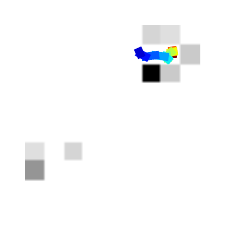}
        \caption{MeRLOT}
    \end{subfigure}
    \caption{
        15-step rollouts with one or three context trajectories. A trajectory starts with red and ends with blue. The walls that can be inferred from context data (i.e. from collisions) are drawn along with the predictions. The darkness of a square reflects the relative number of collisions with each wall.}
    \label{fig:maze_qual}
    \vskip -0.05in
\end{figure}

In meta-testing, we increase the size of the context set to test the meta-scale shift scenario.
Two test cases are considered. First, we measure RMSE of 15-step predictions for 300 different maze configurations. In this test case, we first generate two episodes of 30-length with identical starting positions of the ball. Then, we add additional episodes of the same length that have different initial ball positions than that of query, and measure the 15-step RMSE.
Since most of the additional trajectories are far from a query position, the extra data may hinder inference rather than help. Hence, this test examines the ability of a meta-learner to ignore potentially large amounts of data unrelated to a query.
Figure~\ref{fig:maze_qual} illustrates a test case and sample predicted trajectories.

In the second test case, we directly measure the ability of the learner to infer the underlying maze shape.
This is done by generating four action sequences that can move a ball in four different directions and testing whether the learner predicts if the ball will move in the given direction or not (e.g. whether the learner thinks that the ball hit a wall or not).
This test is done for the cells visited in each trajectory, and the accuracy is measured only for the walls (or absence of walls) that can possibly be inferred from the trajectories.
Similar to the previous test, we measure the accuracy with an increased number of episodes to test meta-scale shift scenario. The accuracy is averaged over 50 different maze configurations. 
\begin{figure}[t]
    \centering
    \begin{subfigure}{0.23\textwidth}
        \includegraphics[width=\textwidth]{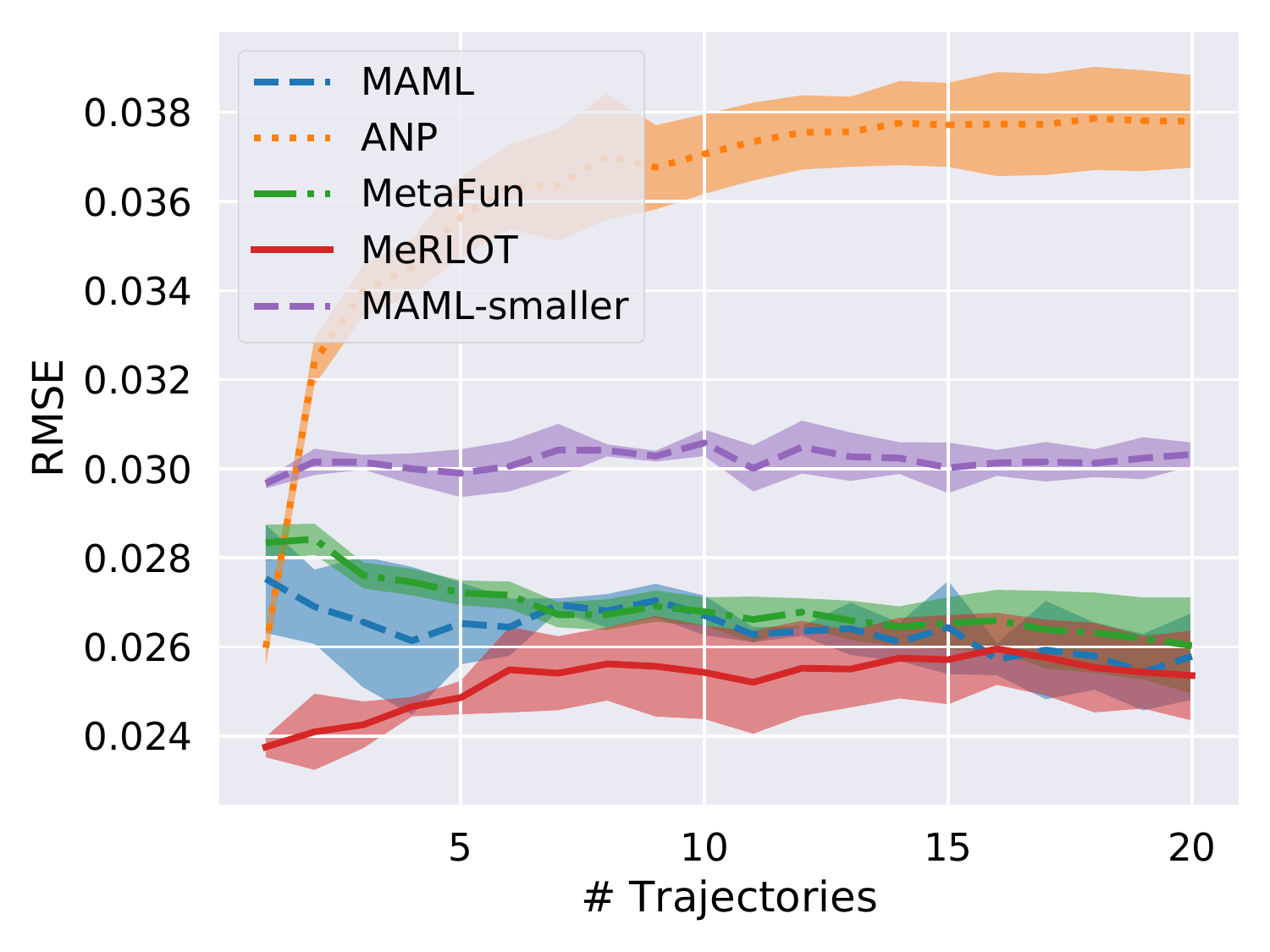}
        \caption{RMSE}
    \end{subfigure}
    \begin{subfigure}{0.23\textwidth}
        \includegraphics[width=\textwidth]{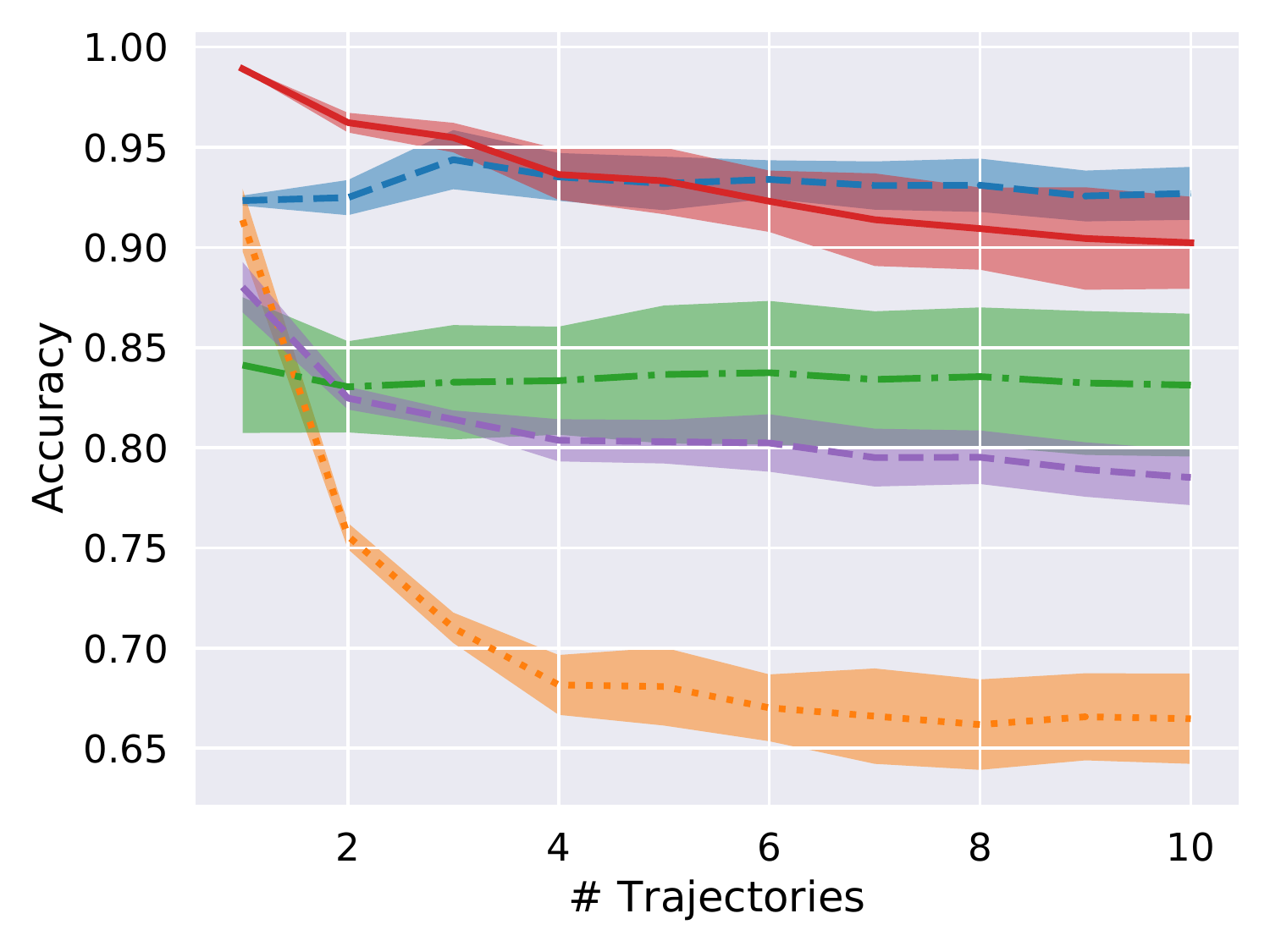}
        \caption{Wall-test accuracy}
    \end{subfigure}
    \caption{Meta-scale shift results on Maze environment. A shaded uncertainty band represents standard deviation of three different runs.}
    \label{fig:maze_scale_change}
    \vskip -0.15in
\end{figure}

The results for the two meta-testing cases are shown in Figure~\ref{fig:maze_scale_change}.
While performance degradation is observed in MeRLOT when more context trajectories are provided, the degradation is more mild compared to that of ANP. At the same time, MeRLOT maintains better performance than MetaFun under both metrics, while MetaFun underfits and shows lower accuracy on the wall prediction test. Notably, MAML achieves good meta-generalization, but only in the case when the the network has a capacity that is well-tuned for the scale of the tasks; when we tried the same experiment with a smaller network (MAML-smaller), meta-generalization ability decreases significantly.
Furthermore, finding this proper network size for MAML can be challenging since MAML takes significantly longer to train than MeRLOT, due to the additional stochastic gradient descent step inside the outer learning loop.

\subsection{Omnipush}
Finally, we validate our approach on the real-world meta-regression dataset, Omnipush \cite{bauza19omnipush}. Omnipush dataset consists of 250 pushes for each of 250 objects having a different shape or mass distribution. Such a large variety of objects results in diverse push behaviors, so the dataset enables us to investigate meta-generalization, especially the meta-range shift scenario.
This dataset is interesting since the problem not only  requires handling discontinuities caused by contact, but the push results also appear to be probabilistic due to unobservable aspects of state.

Meta-generalization is tested following the suggested protocol of the dataset. The input is a three-dimensional vector representing the relative position $x,y$ and the relative angular velocity $\phi$ of the pusher with respect to an object, and the output is a three-dimensional vector $(\Delta x,\Delta y,\Delta \theta)$ representing the change in object pose between the initial position and the final position after a push. The inputs and outputs are normalized across each dimension to have zero mean and standard deviation of 1. For each object, 250 pushes are collected, and 50 of the pushes are provided as a context set and the rest is given as a test set. We report NLL, RMSE, and distance equivalent metric following \citet{bauza19omnipush}.

Four out-of-distribution task sets are suggested in \citet{bauza19omnipush}. First, the 250 objects are split into two groups of 200 and 50 objects, and each of them is used for meta-training and meta-testing.
Second, the pushes of the 10 objects from \citet{Yu2016MoreTA} are considered.
These tasks require more adaptability than the previous tasks since the shapes of the objects do not overlap with the shapes of meta-training objects at all. Therefore, the push behaviors are dissimilar to those of meta-training tasks.
Third, pushes are collected on a different surface having different frictional coefficient. While the pushed objects are included in the meta-training, different friction coefficients can affect outcomes significantly.
Finally, there is a dataset in which the push is collected with the new objects on the new surface.

Table~\ref{tab:omnipush} shows the results of MeRLOT and other approaches. MeRLOT outperforms all other approaches with a large margin with respect to NLL and RMSE except Omnipush Test-set, achieving state-of-the-art in the Omnipush dataset.
MeRLOT especially represents its strength on predicting uncertainty, and it suggests the promise of MeRLOT for applications such as active-learning, in which accurate uncertainty estimation is important.

\begin{table}[t]
\centering
\caption{Quantitative results on Omnipush. OOD task set combines the three out-of-distribution sets. The numbers for MLP and ANP are borrowed from~\protect\citet{bauza19omnipush}.
The experiment is repeated three times with different random seeds, and we reported the average of the runs. Full results with standard deviations are included in the appendix.
}
\vskip 0.1in
\begin{tabular}{llrrr}  
\toprule
Dataset & Method & NLL & RMSE & Dist. Eq. \\
\midrule
\multirow{1}{*}{\begin{tabular}[c]{@{}l@{}}Omnipush\\ Test-set\end{tabular}}
 & MLP & 0.16 & .328 & 7.2 mm \\
 & ANP & -0.11 & \textbf{.225} & 4.9 mm \\
 & MetaFun & 1.59 & .358 & 7.9 mm \\
 & MeRLOT & \textbf{-4.17} & .232 & 5.1 mm \\
\midrule
OOD 
 & MLP & 2.46 & .512 & 11.2 mm \\
 & ANP & 2.33 & .469 & 10.3 mm \\
 & MetaFun & 11.70 & .495 & 10.9 mm \\
 & MeRLOT  & \textbf{-3.21} & \textbf{.425} & 9.3 mm \\
 \midrule
\multirow{1}{*}{\begin{tabular}[c]{@{}l@{}}New\\ surface\end{tabular}}
 & MLP & 1.85 & .333 & 7.3 mm\\
 & ANP & 1.16 & .285 & 6.2 mm \\
 & MetaFun & 11.97 & .361 & 7.9 mm \\
 & MeRLOT  & \textbf{-4.14} & \textbf{.237} & 6.0 mm \\
 \midrule
\multirow{1}{*}{\begin{tabular}[c]{@{}l@{}}New\\ object\end{tabular}}
 & MLP & 2.80 & .601 & 13.2 mm\\
 & ANP & 3.09 & .558 & 12.2 mm\\
 & MetaFun & 22.87 & .610 & 13.4 mm \\
 & MeRLOT  & \textbf{-2.49} & \textbf{.535} & 11.7 mm \\
 \midrule
\multirow{1}{*}{Both} 
 & MLP & 2.72 & .562 & 12.3 mm\\
 & ANP & 2.73 & .517 & 11.3 mm\\
 & MetaFun & 11.56 & .551 & 12.1 mm \\
 & MeRLOT  & \textbf{-2.75} & \textbf{.486} & 10.7 mm \\
\bottomrule
\end{tabular}
\label{tab:omnipush}
\vskip -0.2in
\end{table}

\section{Conclusion}

In this work, we proposed MeRLOT, a novel non-parametric meta-learning algorithm for the problem of meta-regression.
Through a meta-trained local adaptation rule, the algorithm builds a set of functions, each of which is specialized to the neighborhood around its corresponding context data point.
We tested the proposed algorithm for the two types of out-of-distribution tasks and confirmed superior meta-generalization ability for both meta-scale shift and meta-range shift tasks compared to other recent meta-learning algorithms. 
With its improved meta-generalization ability, MeRLOT achieved state-of-the-art results on the real-world robotics dataset, Omnipush.

\section*{Acknowledgements}

This work has taken place in the Personal Autonomous Robotics Lab (PeARL) at The University of Texas at Austin.  PeARL research is supported in part by the NSF (IIS-1724157, IIS-1638107,IIS-1617639, IIS-1749204) and ONR (N00014-18-2243).  
The author would like to thank Ajinkya Jain, Daniel Brown, Jay Whang, and Caleb Chuck for constructive criticism of the manuscript.

\bibliography{bib}

\begin{thebibliography}{30}
\providecommand{\natexlab}[1]{#1}
\providecommand{\url}[1]{\texttt{#1}}
\expandafter\ifx\csname urlstyle\endcsname\relax
  \providecommand{\doi}[1]{doi: #1}\else
  \providecommand{\doi}{doi: \begingroup \urlstyle{rm}\Url}\fi

\bibitem[Bahdanau et~al.(2015)Bahdanau, Cho, and Bengio]{Bahdanau2015NTM}
Bahdanau, D., Cho, K., and Bengio, Y.
\newblock Neural machine translation by jointly learning to align and
  translate.
\newblock In \emph{International Conference on Learning Representations}, 2015.

\bibitem[Bauza et~al.(2019)Bauza, Alet, Lin, Lozano-Perez, Kaelbling, Isola,
  and Rodriguez]{bauza19omnipush}
Bauza, M., Alet, F., Lin, Y.-C., Lozano-Perez, T., Kaelbling, L.~P., Isola, P.,
  and Rodriguez, A.
\newblock Omnipush: accurate, diverse, real-world dataset of pushing dynamics
  with rgb-d video.
\newblock In \emph{International Conference on Intelligent Robots and Systems},
  2019.

\bibitem[Clavera et~al.(2019)Clavera, Nagabandi, Liu, Fearing, Abbeel, Levine,
  and Finn]{clavera2018learning}
Clavera, I., Nagabandi, A., Liu, S., Fearing, R.~S., Abbeel, P., Levine, S.,
  and Finn, C.
\newblock Learning to adapt in dynamic, real-world environments through
  meta-reinforcement learning.
\newblock In \emph{International Conference on Learning Representations}, 2019.

\bibitem[Duan et~al.(2016)Duan, Schulman, Chen, Bartlett, Sutskever, and
  Abbeel]{duan2016rl2}
Duan, Y., Schulman, J., Chen, X., Bartlett, P.~L., Sutskever, I., and Abbeel,
  P.
\newblock Rl{$^2$}: Fast reinforcement learning via slow reinforcement
  learning.
\newblock \emph{arXiv preprint arXiv:1611.02779}, 2016.

\bibitem[Finn \& Levine(2018)Finn and Levine]{finn18UnivMAML}
Finn, C. and Levine, S.
\newblock Meta-learning and universality: Deep representations and gradient
  descent can approximate any learning algorithm.
\newblock In \emph{International Conference on Learning Representations}, 2018.

\bibitem[Finn et~al.(2017)Finn, Abbeel, and Levine]{finn2017MAML}
Finn, C., Abbeel, P., and Levine, S.
\newblock Model-agnostic meta-learning for fast adaptation of deep networks.
\newblock In \emph{Proceedings of the 34th International Conference on Machine
  Learning-Volume 70}, pp.\  1126--1135, 2017.

\bibitem[Finn et~al.(2018)Finn, Xu, and Levine]{finn18probMAML}
Finn, C., Xu, K., and Levine, S.
\newblock Probabilistic model-agnostic meta-learning.
\newblock In \emph{Advances in Neural Information Processing Systems 31}, pp.\
  9516--9527, 2018.

\bibitem[Garnelo et~al.(2018)Garnelo, Rosenbaum, Maddison, Ramalho, Saxton,
  Shanahan, Teh, Rezende, and Eslami]{garnelo18cnp}
Garnelo, M., Rosenbaum, D., Maddison, C., Ramalho, T., Saxton, D., Shanahan,
  M., Teh, Y.~W., Rezende, D., and Eslami, S. M.~A.
\newblock Conditional neural processes.
\newblock In \emph{Proceedings of the 35th International Conference on Machine
  Learning-Volume 80}, pp.\  1704--1713, 2018.

\bibitem[Grant et~al.(2018)Grant, Finn, Levine, Darrell, and
  Griffiths]{grant2018recasting}
Grant, E., Finn, C., Levine, S., Darrell, T., and Griffiths, T.
\newblock Recasting gradient-based meta-learning as hierarchical bayes.
\newblock In \emph{International Conference on Learning Representations}, 2018.

\bibitem[Guiroy et~al.(2019)Guiroy, Verma, and Pal]{guiroy2019understanding}
Guiroy, S., Verma, V., and Pal, C.
\newblock Towards understanding generalization in gradient-based meta-learning.
\newblock \emph{arXiv preprint arXiv:1907.07287}, 2019.

\bibitem[Jamal \& Qi(2019)Jamal and Qi]{Jamal2019TAML}
Jamal, M.~A. and Qi, G.-J.
\newblock Task agnostic meta-learning for few-shot learning.
\newblock In \emph{Proceedings of the IEEE Conference on Computer Vision and
  Pattern Recognition}, pp.\  11719--11727, 2019.

\bibitem[Kim et~al.(2019)Kim, Mnih, Schwarz, Garnelo, Eslami, Rosenbaum,
  Vinyals, and Teh]{kim18anp}
Kim, H., Mnih, A., Schwarz, J., Garnelo, M., Eslami, A., Rosenbaum, D.,
  Vinyals, O., and Teh, Y.~W.
\newblock Attentive neural processes.
\newblock In \emph{International Conference on Learning Representations}, 2019.

\bibitem[Lee \& Choi(2018)Lee and Choi]{lee2018MTNet}
Lee, Y. and Choi, S.
\newblock Gradient-based meta-learning with learned layerwise metric and
  subspace.
\newblock In \emph{International Conference on Machine Learning}, pp.\
  2933--2942, 2018.

\bibitem[Mason et~al.(1999)Mason, Baxter, Bartlett, and
  Frean]{manson2000functional}
Mason, L., Baxter, J., Bartlett, P., and Frean, M.
\newblock Functional gradient techniques for combining hypotheses.
\newblock \emph{Advances in Large Margin Classifiers}, 1999.

\bibitem[Mishra et~al.(2018)Mishra, Rohaninejad, Chen, and
  Abbeel]{mishra2018SNAIL}
Mishra, N., Rohaninejad, M., Chen, X., and Abbeel, P.
\newblock A simple neural attentive meta-learner.
\newblock In \emph{International Conference on Learning Representations}, 2018.

\bibitem[Na et~al.(2020)Na, Lee, Lee, Kim, Park, Yang, and Hwang]{na2019ood}
Na, D., Lee, H.~B., Lee, H., Kim, S., Park, M., Yang, E., and Hwang, S.~J.
\newblock Learning to balance: Bayesian meta-learning for imbalanced and
  out-of-distribution tasks.
\newblock In \emph{International Conference on Learning Representations}, 2020.

\bibitem[Ravi \& Larochelle(2017)Ravi and Larochelle]{Ravi2017MetaLSTM}
Ravi, S. and Larochelle, H.
\newblock Optimization as a model for few-shot learning.
\newblock In \emph{In International Conference on Learning Representations},
  2017.

\bibitem[Santoro et~al.(2016)Santoro, Bartunov, Botvinick, Wierstra, and
  Lillicrap]{santoro16MANN}
Santoro, A., Bartunov, S., Botvinick, M., Wierstra, D., and Lillicrap, T.
\newblock Meta-learning with memory-augmented neural networks.
\newblock In \emph{Proceedings of the 33rd International Conference on
  International Conference on Machine Learning - Volume 48}, pp.\  1842–1850,
  2016.

\bibitem[Schmidhuber(1987)]{schmidhuber1987meta-learning}
Schmidhuber, J.
\newblock \emph{Evolutionary principles in self-referential learning, or on
  learning how to learn: The meta-meta-... hook}.
\newblock Diplomarbeit, Technische Universität München, München, 1987.

\bibitem[Snell et~al.(2017)Snell, Swersky, and Zemel]{snell2017prototypical}
Snell, J., Swersky, K., and Zemel, R.
\newblock Prototypical networks for few-shot learning.
\newblock In \emph{Advances in Neural Information Processing Systems 30}, pp.\
  4077--4087, 2017.

\bibitem[Thrun \& Pratt(2012)Thrun and Pratt]{thrun2012learning}
Thrun, S. and Pratt, L.
\newblock \emph{Learning to learn}.
\newblock Springer Science \& Business Media, 2012.

\bibitem[Tossou et~al.(2019)Tossou, Dura, Laviolette, Marchand, and
  Lacoste]{tossou2019adaDKL}
Tossou, P., Dura, B., Laviolette, F., Marchand, M., and Lacoste, A.
\newblock Adaptive deep kernel learning.
\newblock \emph{arXiv preprint arXiv:1905.12131}, 2019.

\bibitem[Vaswani et~al.(2017)Vaswani, Shazeer, Parmar, Uszkoreit, Jones, Gomez,
  Kaiser, and Polosukhin]{vaswani17Attention}
Vaswani, A., Shazeer, N., Parmar, N., Uszkoreit, J., Jones, L., Gomez, A.~N.,
  Kaiser, u., and Polosukhin, I.
\newblock Attention is all you need.
\newblock In \emph{Proceedings of the 31st International Conference on Neural
  Information Processing Systems}, pp.\  6000–6010, 2017.

\bibitem[Vinyals et~al.(2016)Vinyals, Blundell, Lillicrap, Kavukcuoglu, and
  Wierstra]{Vinyals16MatchingNet}
Vinyals, O., Blundell, C., Lillicrap, T., Kavukcuoglu, K., and Wierstra, D.
\newblock Matching networks for one shot learning.
\newblock In \emph{Proceedings of the 30th International Conference on Neural
  Information Processing Systems}, pp.\  3637–3645, 2016.

\bibitem[Wang et~al.(2016)Wang, Kurth-Nelson, Soyer, Leibo, Tirumala, Munos,
  Blundell, Kumaran, and Botvinick]{Wang2016LearningTR}
Wang, J.~X., Kurth-Nelson, Z., Soyer, H., Leibo, J.~Z., Tirumala, D., Munos,
  R., Blundell, C., Kumaran, D., and Botvinick, M.~M.
\newblock Learning to reinforcement learn.
\newblock \emph{arXiv preprint arXiv:1611.05763}, 2016.

\bibitem[Xu et~al.(2019)Xu, Ton, Kim, Kosiorek, and Teh]{xu2019metafun}
Xu, J., Ton, J.-F., Kim, H., Kosiorek, A.~R., and Teh, Y.~W.
\newblock Metafun: Meta-learning with iterative functional updates.
\newblock \emph{arXiv preprint arXiv:1912.02738}, 2019.

\bibitem[Yin et~al.(2020)Yin, Tucker, Zhou, Levine, and
  Finn]{yin2020metaMemorization}
Yin, M., Tucker, G., Zhou, M., Levine, S., and Finn, C.
\newblock Meta-learning without memorization.
\newblock In \emph{International Conference on Learning Representations}, 2020.

\bibitem[Yoon et~al.(2018)Yoon, Kim, Dia, Kim, Bengio, and
  Ahn]{taesup18BayesianMAML}
Yoon, J., Kim, T., Dia, O., Kim, S., Bengio, Y., and Ahn, S.
\newblock Bayesian model-agnostic meta-learning.
\newblock In \emph{Advances in Neural Information Processing Systems 31}, pp.\
  7332--7342, 2018.

\bibitem[Yu et~al.(2016)Yu, Bauz{\'a}, Fazeli, and Rodr{\'i}guez]{Yu2016MoreTA}
Yu, K.-T., Bauz{\'a}, M., Fazeli, N., and Rodr{\'i}guez, A.
\newblock More than a million ways to be pushed. a high-fidelity experimental
  dataset of planar pushing.
\newblock \emph{IEEE/RSJ International Conference on Intelligent Robots and
  Systems}, pp.\  30--37, 2016.

\bibitem[Zintgraf et~al.(2019)Zintgraf, Shiarli, Kurin, Hofmann, and
  Whiteson]{zintgraf2018CAVIA}
Zintgraf, L., Shiarli, K., Kurin, V., Hofmann, K., and Whiteson, S.
\newblock Fast context adaptation via meta-learning.
\newblock In \emph{Proceedings of the 36th International Conference on Machine
  Learning}, pp.\  7693--7702, 2019.

\end{thebibliography}
\bibliographystyle{icml2020}

\onecolumn
\newpage
\appendix
\setcounter{figure}{0} \renewcommand{\thefigure}{A.\arabic{figure}}
\setcounter{table}{0} \renewcommand{\thetable}{A.\arabic{table}}
\setcounter{footnote}{0}
\section{Experimental Details}

All the models are trained with GTX 1080 Ti, Titan X Pascal, Titan V, or Titan V100 GPUs. We implemented the code with Python 3.7 and Tensorflow v2.1.0, and we used the default hyperparameters of Tensorflow unless noted otherwise.
The model architectures and hyperparameters are displayed in Table~\ref{tab:hyperparams}. 
We performed only a minor hyperparameter search (roughly 10 total settings), until values were found that stabilized the learning progress of MeRLOT on 1D function regression tasks, and we used the same hyperparameters for all the other experiments. We considered different inner and outer learning rates, the width of the fully connected layer $d_r$, and the number of the fully connected layer $K$. We found that a higher outer learning rate could introduce instability in training.
For all the three experiments, we used $d_r = 128$ and $K=2$ except MAML where we used $d_r=512$ and $K=5$, which resulted in a network that has more parameters than the other three methods. For the smaller MAML used in the maze experiments, we used $d_r = 128$ and $K=2$.
For the 1D function regression problem and Omnipush, in which each model predicts a Gaussian distribution instead of a point estimate, we modeled mean and standard deviation. For the 1D function regression problem, standard deviation is modeled as $\sigma = 0.1 + 0.9 \times \text{softplus}(x)$, and for Omnipush, standard deviation is modeled as $\sigma = \text{softplus}(x)$, following the ANP implementation of \citet{bauza19omnipush}.

\begin{table}[ht]
\caption{
The model architecture and hyperparameters used in the experiment. FC($a,b\times K,c$) represents a $K$+1-layer fully connected network where the input and output has a dimension $a$ and $b$ while the network has $b$ units inside. Each fully connected network has \texttt{relu} nonlinearity in between except the final layer. $d_x$, $d_y$, and $d_r$ represent the dimension of input $x$, output $y$ and functional representation $r$.
}
\vspace{0.1in}
\centering
\begin{tabular}{rcccc}
\toprule
 & MAML & ANP & MetaFun & MeRLOT \\
\midrule
\# Adaptation ($T$) & 3 & N/A & 3 & 3 \\
Inner lr ($\alpha$) & 0.01 & N/A & 0.01 & 0.01 \\
Outer lr & 1e-3 & 5e-5 & 5e-5 & 5e-5 \\
$x$ encoder & N/A & FC($d_x$,$d_r$$\times$K,$d_r$) & FC($d_x$,$d_r$$\times$K,$d_r$) & FC($d_x$,$d_r$$\times$K,$d_r$) \\
$(x,y)$ encoder & N/A & 2 $\times$ FC($d_x$+$d_y$,$d_r$$\times$K,$d_r$) \footnotemark & N/A & FC($d_x$+$d_y$,$d_r$$\times$K,$d_r$) \\
\# attention layers & N/A & 2 $\times$ K & N/A & K \\
\# attention heads & N/A & 8 & N/A & 8 \\
Latent $z$ encoder & N/A & FC($d_r$,$d_r \times 1$,$d_r$) & N/A & N/A \\
$\psi$ & N/A & N/A & N/A & FC($d_x$+$d_y$,$d_r$$\times$K,$d_r$) \\
$r_x^i(0)$ & N/A & N/A & N/A & FC($d_x$+$d_r$,$d_r$$\times$K,$d_r$) \\
Updater & N/A & N/A & \multicolumn{2}{c}{FC($d_x$+$d_y$+$d_r$,$d_r$$\times$K,$d_r$)} \\
Decoder & FC($d_x$+$d_y$,$d_r$$\times$K,2$\cdot d_y$) & FC($d_x$+2$\cdot d_r$,$d_r$$\times$K,2$\cdot d_y$) & \multicolumn{2}{c}{FC($d_x$+$d_r$,$d_r$$\times$K,2$\cdot d_y$)} \\
 \midrule
 \# Parameters &
 $\sim$ 1M & $\sim$ 520K & $\sim$ 110K & $\sim$ 760K \\
\bottomrule
\end{tabular}
\label{tab:hyperparams}
\end{table}
\footnotetext{ANP has two pairs of $(x,y)$ encoder and attention layers since it has the query dependent deterministic path for $r$ and the latent path for $z$. For a detailed architecture, please refer the original paper.}

\section{Supplementary Results}

\begin{figure*}[h]
    \centering
    \begin{subfigure}{0.245\textwidth}
        \includegraphics[width=\textwidth]{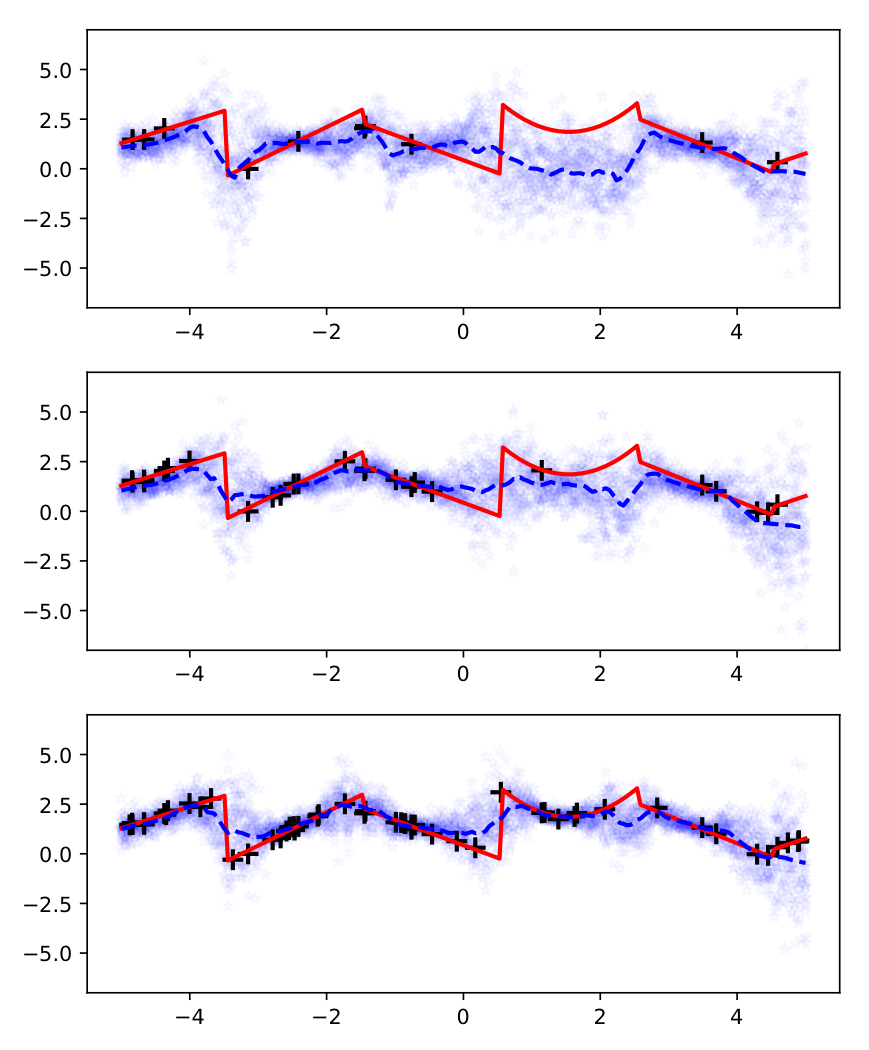}
        \caption{MAML}
    \end{subfigure}
    \begin{subfigure}{0.245\textwidth}
        \includegraphics[width=\textwidth]{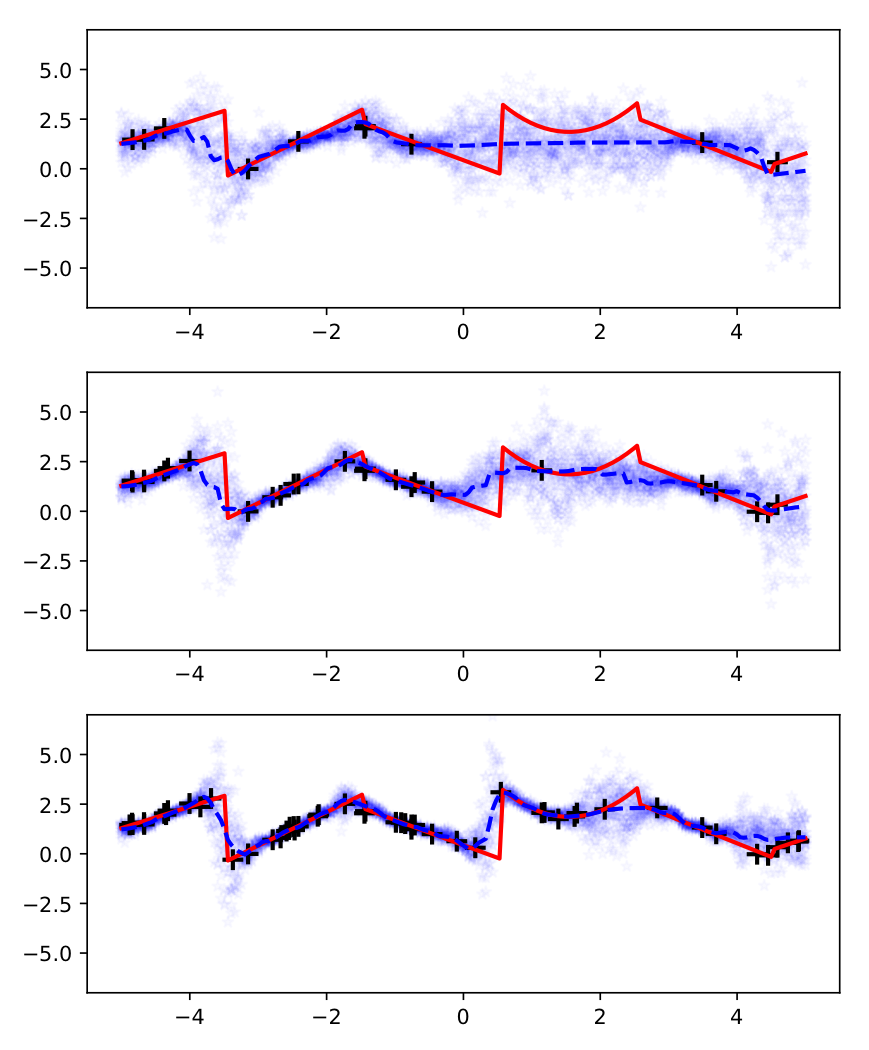}
        \caption{ANP}
    \end{subfigure}
    \begin{subfigure}{0.245\textwidth}
        \includegraphics[width=\textwidth]{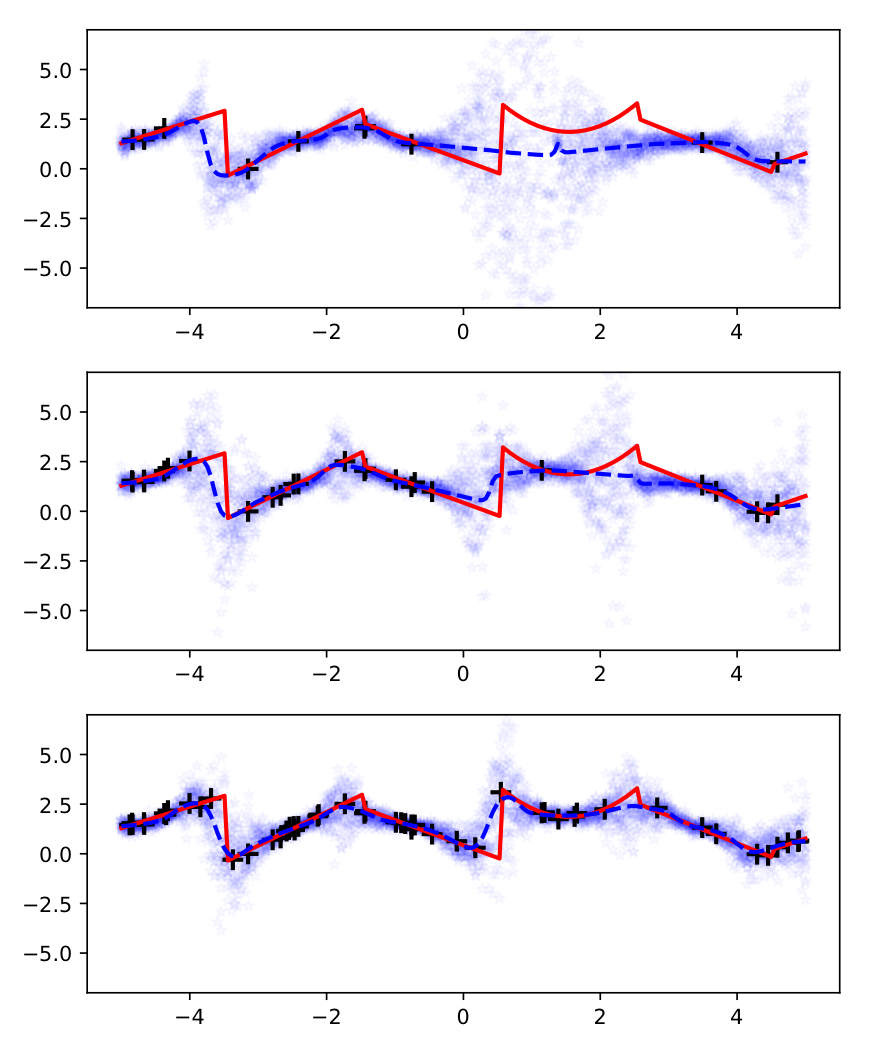}
        \caption{MetaFun}
    \end{subfigure}
    \begin{subfigure}{0.245\textwidth}
        \includegraphics[width=\textwidth]{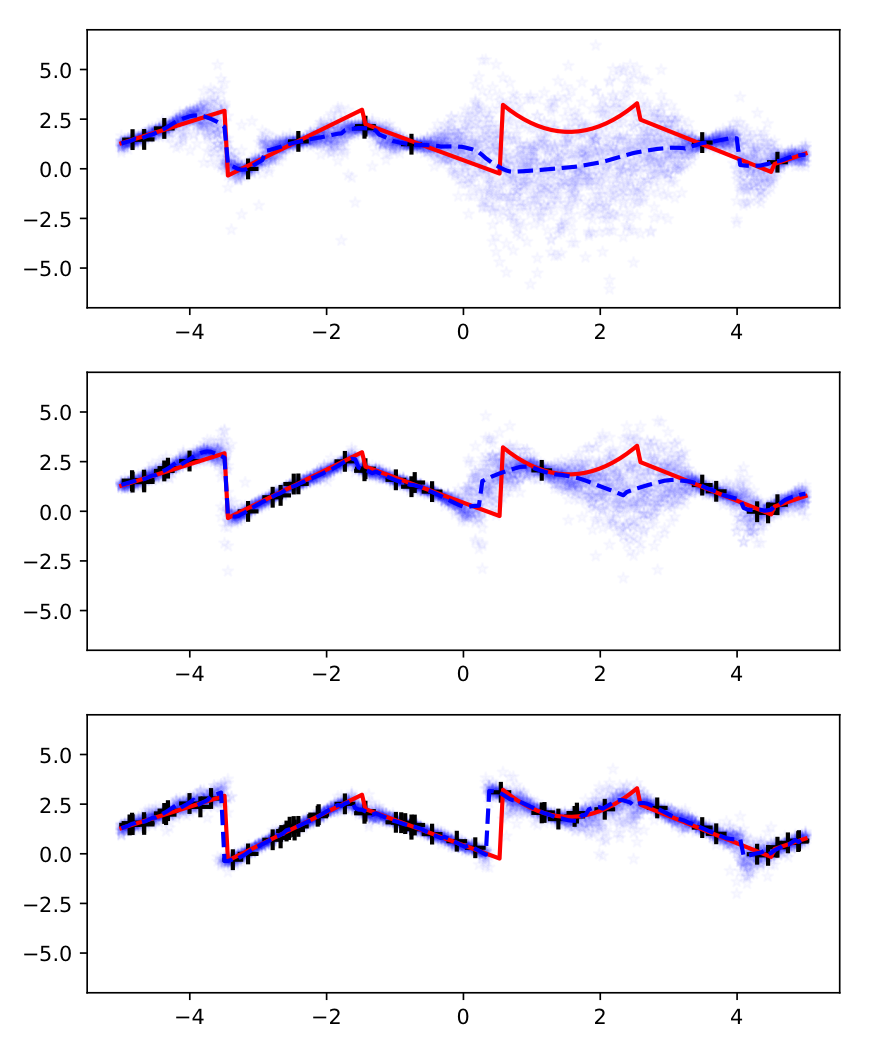}
        \caption{MeRLOT}
    \end{subfigure}
    
    \begin{subfigure}{0.245\textwidth}
        \includegraphics[width=\textwidth]{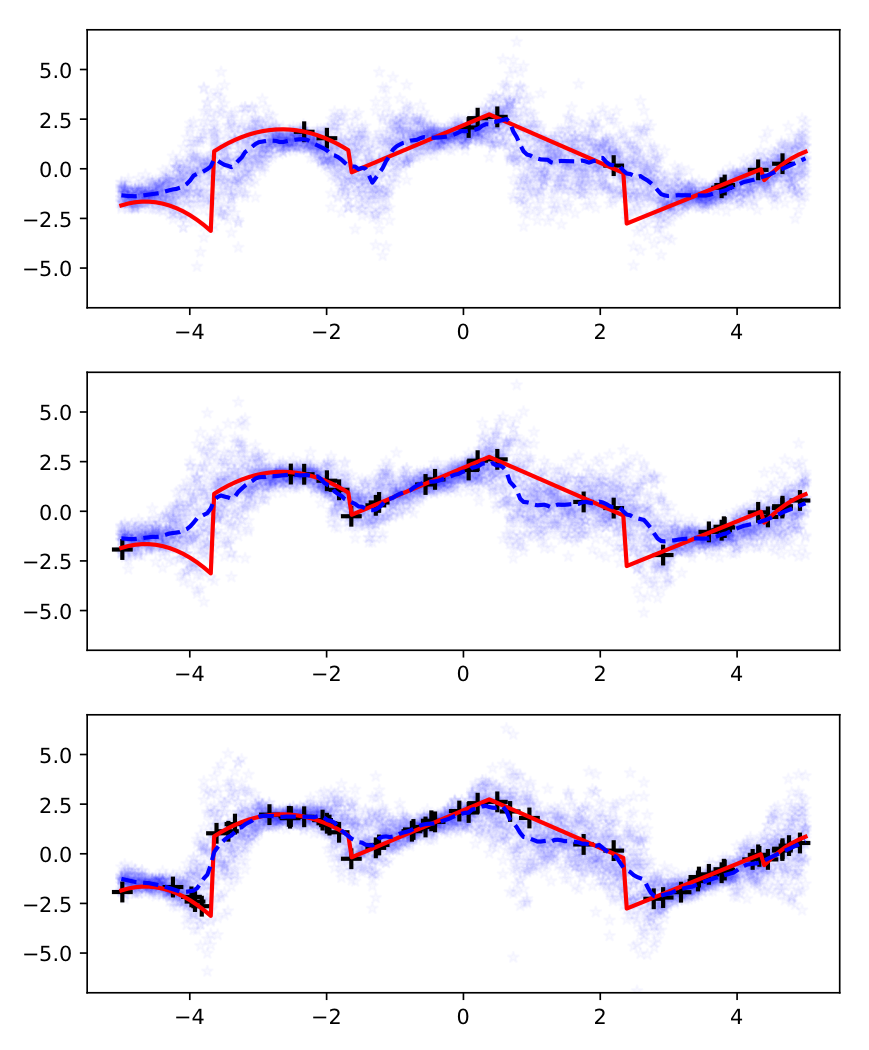}
        \caption{MAML}
    \end{subfigure}
    \begin{subfigure}{0.245\textwidth}
        \includegraphics[width=\textwidth]{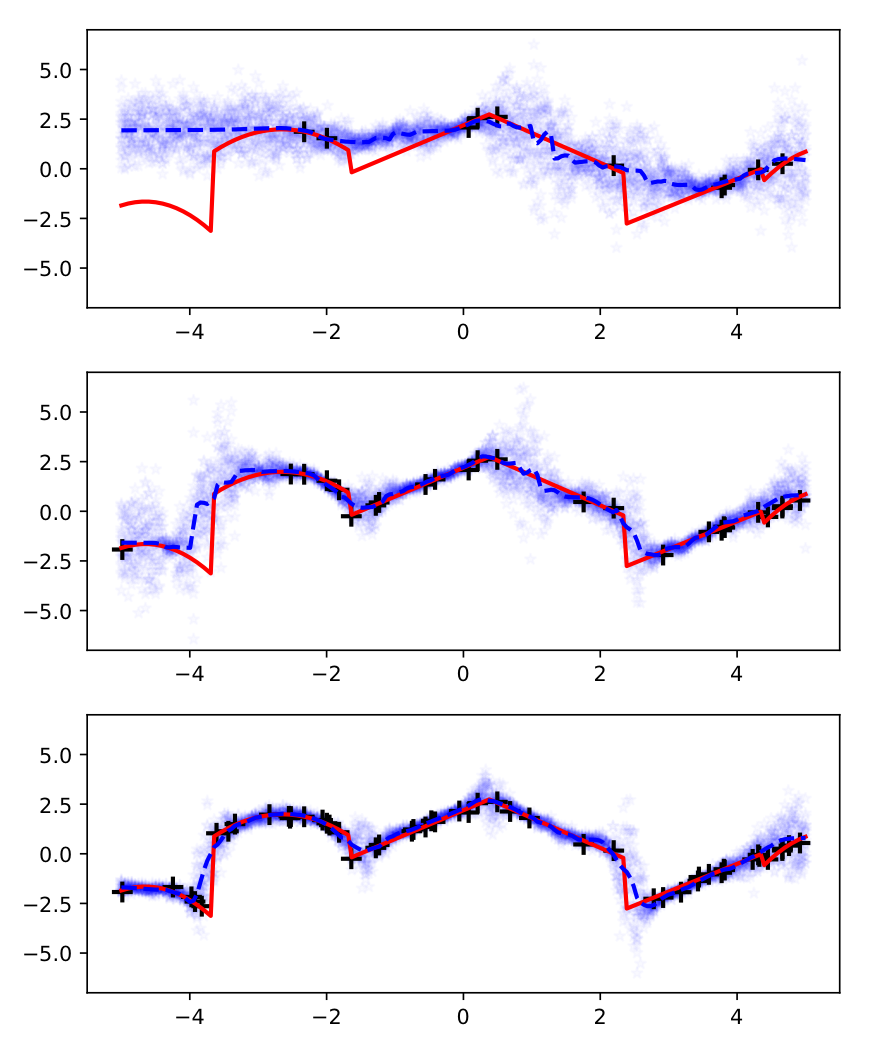}
        \caption{ANP}
    \end{subfigure}
    \begin{subfigure}{0.245\textwidth}
        \includegraphics[width=\textwidth]{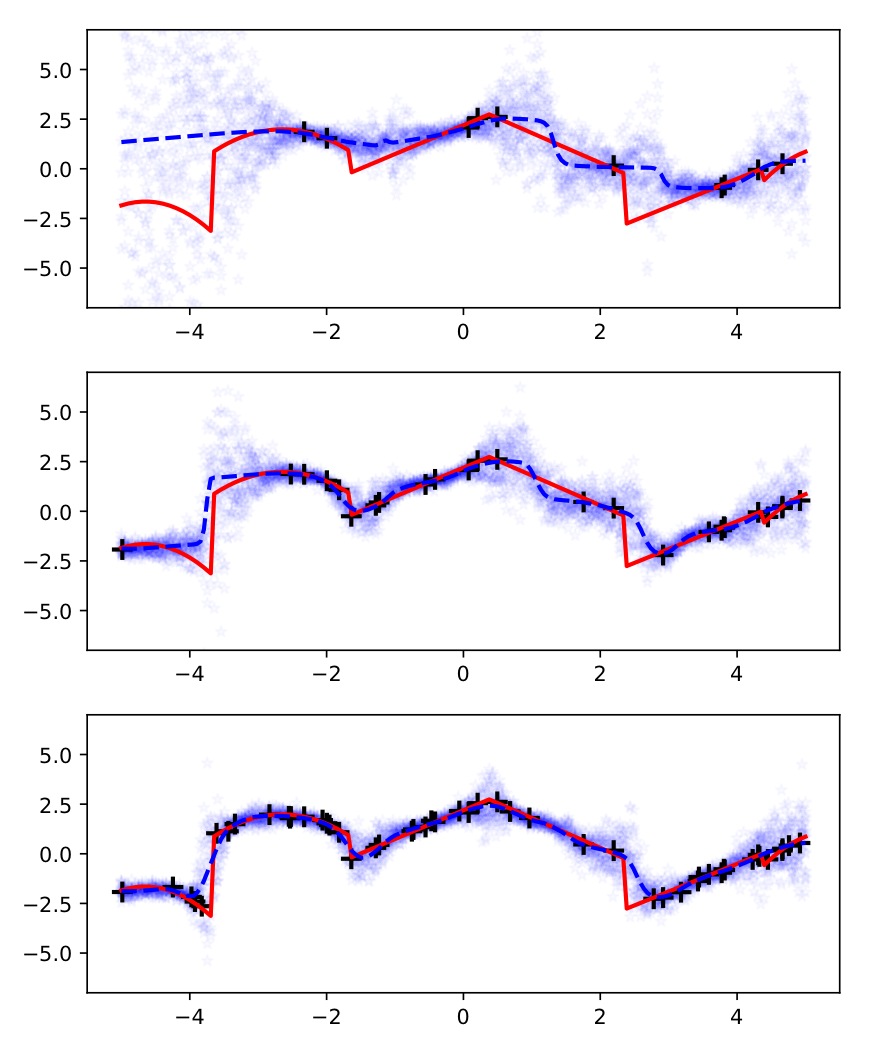}
        \caption{MetaFun}
    \end{subfigure}
    \begin{subfigure}{0.245\textwidth}
        \includegraphics[width=\textwidth]{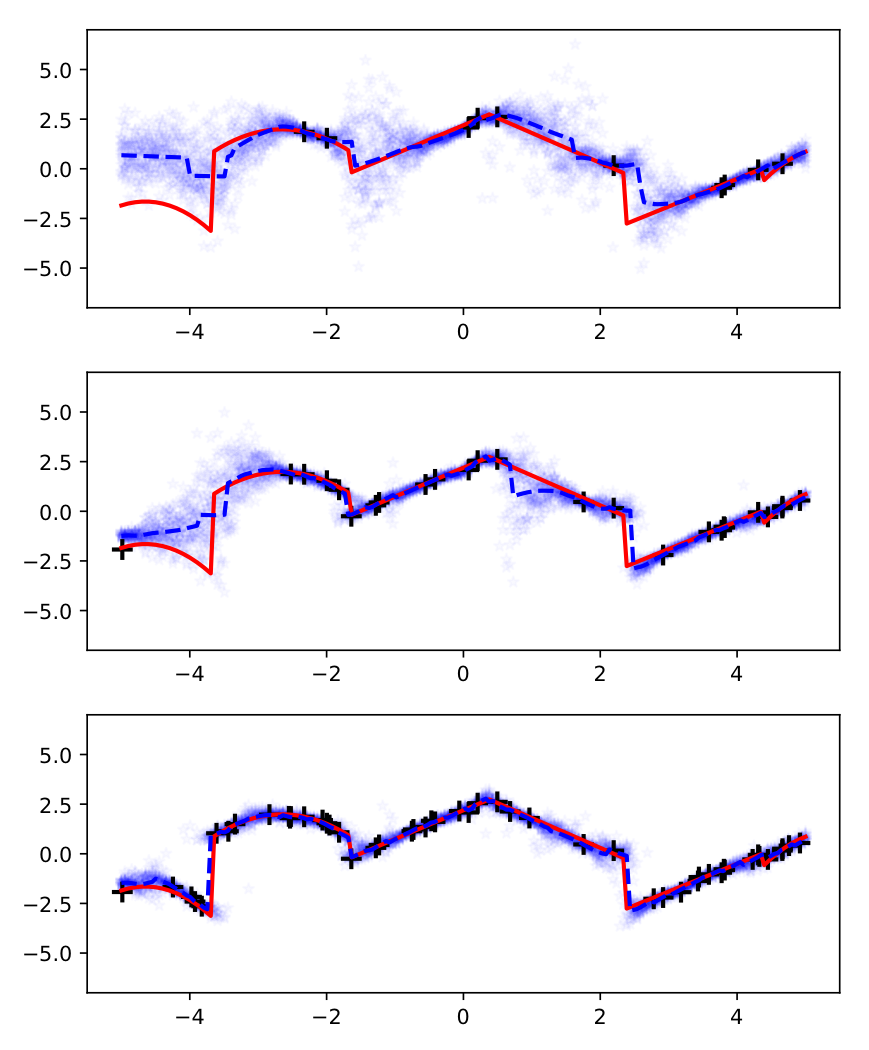}
        \caption{MeRLOT}
    \end{subfigure}
    
    \begin{subfigure}{0.245\textwidth}
        \includegraphics[width=\textwidth]{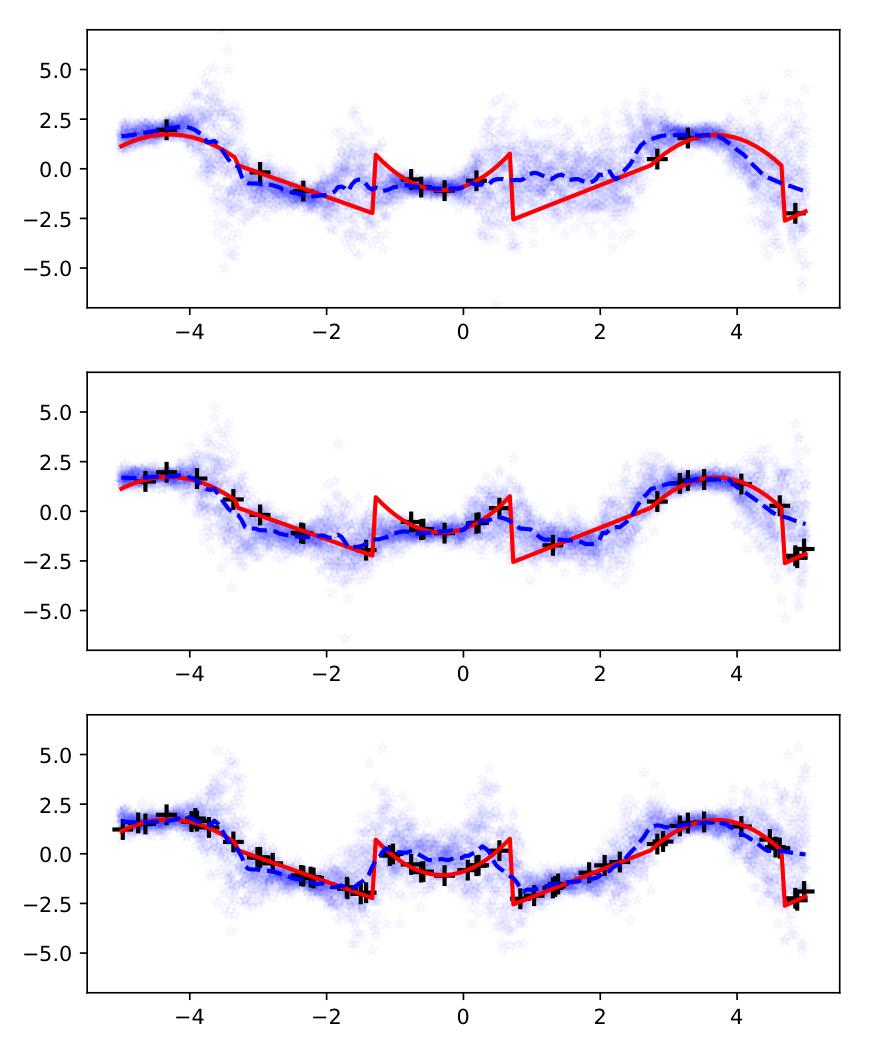}
        \caption{MAML}
    \end{subfigure}
    \begin{subfigure}{0.245\textwidth}
        \includegraphics[width=\textwidth]{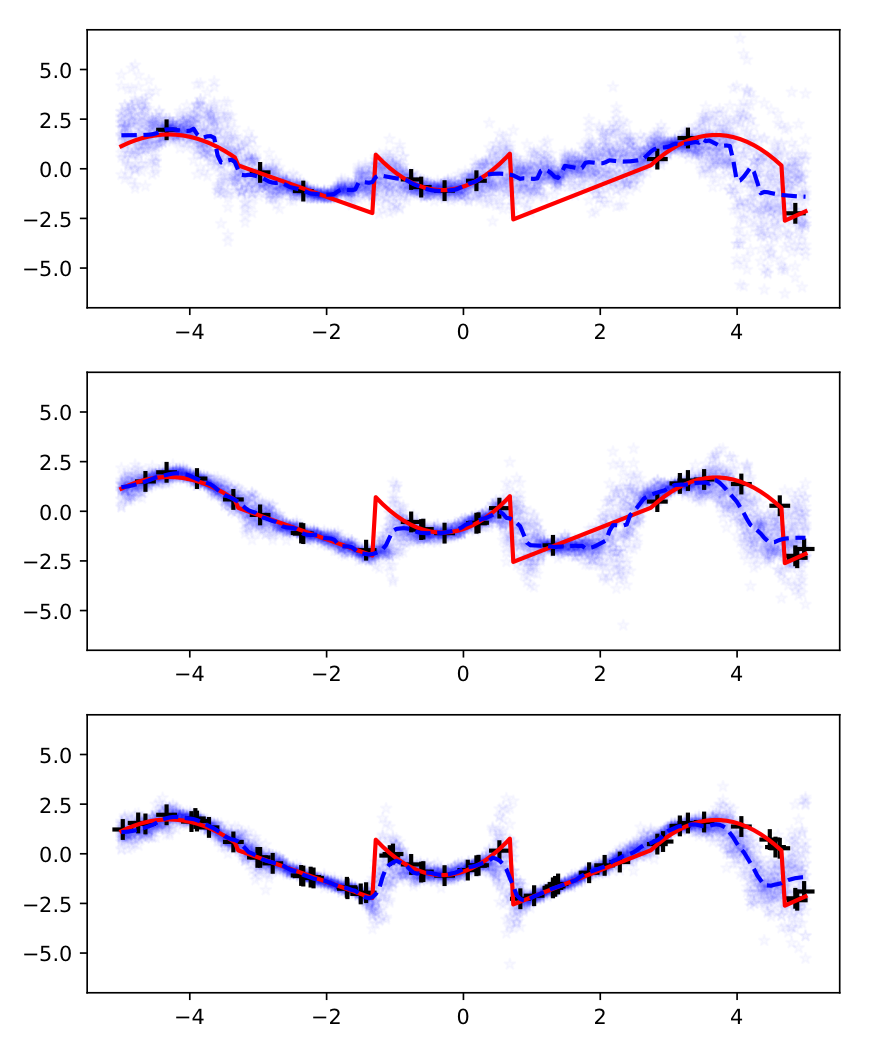}
        \caption{ANP}
    \end{subfigure}
    \begin{subfigure}{0.245\textwidth}
        \includegraphics[width=\textwidth]{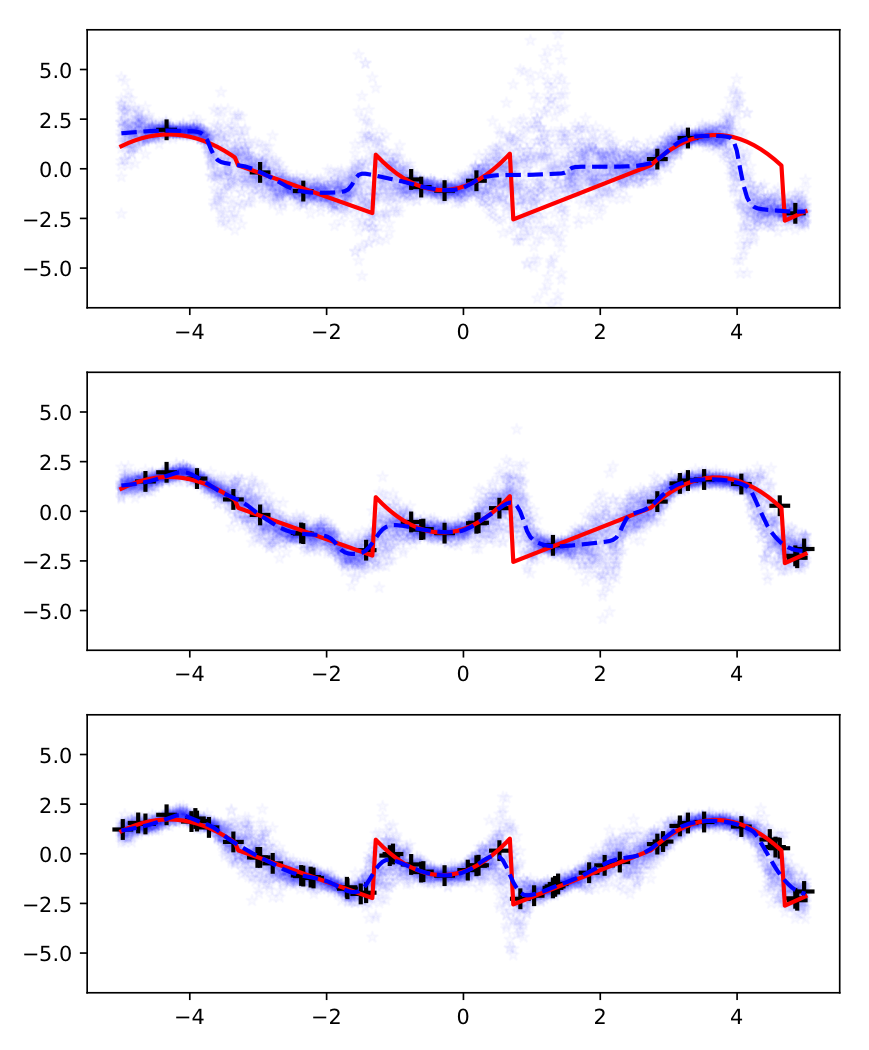}
        \caption{MetaFun}
    \end{subfigure}
    \begin{subfigure}{0.245\textwidth}
        \includegraphics[width=\textwidth]{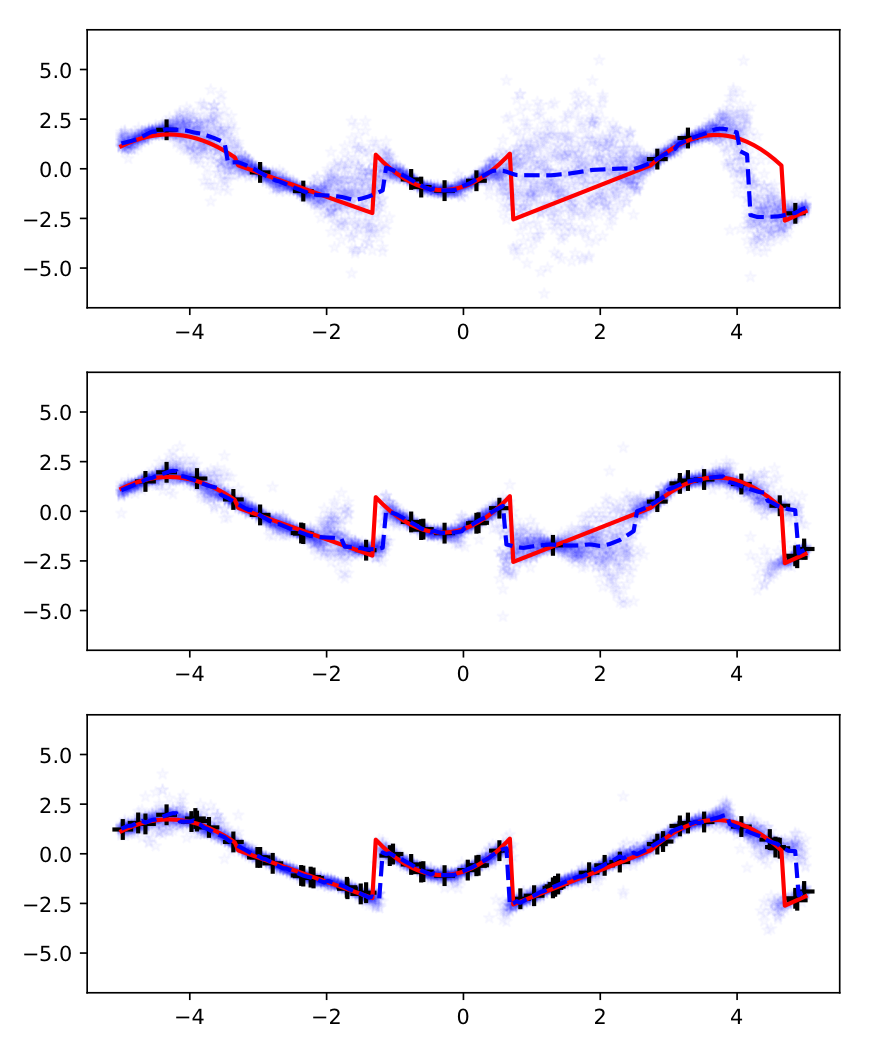}
        \caption{MeRLOT}
    \end{subfigure}
    \caption{More qualitative results of meta-scale shift on 1D function regression}
\end{figure*}

\begin{figure*}[t]
    \centering
    \begin{subfigure}{\textwidth}
        \includegraphics[width=\textwidth]{Figures/locality_test_ours.pdf}
        \caption{MeRLOT}
        \label{fig:locality:full}
    \end{subfigure}
    \begin{subfigure}{\textwidth}
        \includegraphics[width=\textwidth]{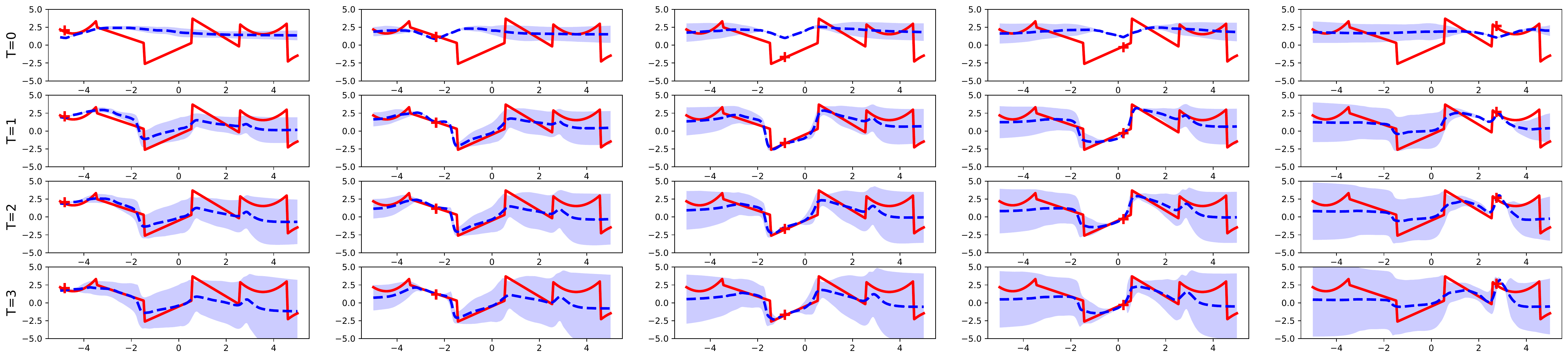}
        \caption{MeRLOT w/o the seed function generator $\psi$}
        \label{fig:locality:ablated}
    \end{subfigure}
    \caption{
    Visualization of local adaptation with and without the seed function generator $\psi$.
    Each row represents different update iterations $t$, and each column represents the different local function $f^i(t)$ based on each context data point $(x_i,y_i)$. Each context point is drawn with a red plus symbol $+$.
    The mean of $f^i(t)$ is drawn as a blue line with a shaded uncertainty band representing standard deviation of the prediction.
    We used a meta-scale shift set with context size of 50, and we choose 5 (out of 50) local functions to display.
    The modifications on each function made by the updater $u$ are more locally constrained when there is a seed function generator $\psi$, which generates initial functions for each given context data point.
    }
\end{figure*}

\begin{table*}[h]
\centering
\caption{Quantitative results for two out-of-distribution meta-test scenarios for 1D function regression. The experiment is repeated three times with different random seeds, and we reported the average of the runs with standard deviation. The top rows of the table display results for baselines and MeRLOT; lower rows show ablations of MetaFun and MeRLOT with self attention added and seed generation removed, respectively. For ease of comparison, the left side of the table displays whether each method includes self attention, iterative updates, local functions, and seed generation ($\psi$).}
\vspace{0.1in}
\begin{tabular}{lccccrrrrrrrr}
\toprule
 & \multirow{3}{*}{\begin{tabular}[c]{@{}c@{}}Self\\ Att.\end{tabular}} & \multirow{3}{*}{\begin{tabular}[c]{@{}c@{}}Iter.\\ Upd.\end{tabular}} & \multirow{3}{*}{\begin{tabular}[c]{@{}c@{}}Loc.\\ Fns.\end{tabular}} & \multirow{3}{*}{$\psi$} & \multicolumn{4}{l}{Meta-range shift} & \multicolumn{4}{l}{Meta-scale shift} \\
Method & & & & & \multicolumn{2}{l}{Interpolation} & \multicolumn{2}{l}{Extrapolation} & \multicolumn{2}{l}{10 shots} & \multicolumn{2}{l}{50 shots} \\ 
\multicolumn{1}{c}{} &  &  &  &  & NLL & RMSE & NLL & RMSE & NLL & RMSE & NLL & RMSE \\
\midrule
MAML & \multicolumn{4}{c}{N/A} & \begin{tabular}[c]{@{}r@{}}0.828\\ (0.051)\end{tabular} & \begin{tabular}[c]{@{}r@{}}0.528\\ (0.016)\end{tabular} & \begin{tabular}[c]{@{}r@{}}1.806\\ (0.139)\end{tabular} & \begin{tabular}[c]{@{}r@{}}1.686\\ (0.094)\end{tabular} & \begin{tabular}[c]{@{}r@{}}2.127\\ (0.096)\end{tabular} & \begin{tabular}[c]{@{}r@{}}1.314\\ (0.023)\end{tabular} & \begin{tabular}[c]{@{}r@{}}1.068\\ (0.060)\end{tabular} & \begin{tabular}[c]{@{}r@{}}1.019\\ (0.035)\end{tabular} \\
ANP & \cmark & \xmark & \multicolumn{2}{c}{N/A} & \begin{tabular}[c]{@{}r@{}}0.883\\ (0.088)\end{tabular} & \begin{tabular}[c]{@{}r@{}}0.425\\ (0.027)\end{tabular} & \begin{tabular}[c]{@{}r@{}}1.920\\ (0.133)\end{tabular} & \begin{tabular}[c]{@{}r@{}}1.331\\ (0.020)\end{tabular} & \begin{tabular}[c]{@{}r@{}}1.970\\ (0.077)\end{tabular} & \begin{tabular}[c]{@{}r@{}}1.167\\ (0.010)\end{tabular} & \begin{tabular}[c]{@{}r@{}}0.093\\ (0.097)\end{tabular} & \begin{tabular}[c]{@{}r@{}}0.585\\ (0.037)\end{tabular} \\
MetaFun & \xmark & \cmark & \xmark & \xmark & \begin{tabular}[c]{@{}r@{}}0.524\\ (0.006)\end{tabular} & \begin{tabular}[c]{@{}r@{}}0.316\\ (0.001)\end{tabular} & \begin{tabular}[c]{@{}r@{}}1.563\\ (0.126)\end{tabular} & \begin{tabular}[c]{@{}r@{}}1.297\\ (0.019)\end{tabular} & \begin{tabular}[c]{@{}r@{}}1.433\\ (0.083)\end{tabular} & \begin{tabular}[c]{@{}r@{}}1.230\\ (0.012)\end{tabular} & \begin{tabular}[c]{@{}r@{}}0.199\\ (0.065)\end{tabular} & \begin{tabular}[c]{@{}r@{}}0.614\\ (0.006)\end{tabular} \\
MeRLOT & \cmark & \cmark & \cmark & \cmark & \begin{tabular}[c]{@{}r@{}}-0.059\\ (0.014)\end{tabular} & \begin{tabular}[c]{@{}r@{}}0.331\\ (0.002)\end{tabular} & \begin{tabular}[c]{@{}r@{}}1.104\\ (0.116)\end{tabular} & \begin{tabular}[c]{@{}r@{}}1.143\\ (0.023)\end{tabular} & \begin{tabular}[c]{@{}r@{}}0.896\\ (0.075)\end{tabular} & \begin{tabular}[c]{@{}r@{}}1.121\\ (0.028)\end{tabular} & \begin{tabular}[c]{@{}r@{}}-0.453\\ (0.086)\end{tabular} & \begin{tabular}[c]{@{}r@{}}0.570\\ (0.025)\end{tabular} \\
\midrule
\begin{tabular}[c]{@{}l@{}}MetaFun\\ w/ SA\end{tabular} & \cmark & \cmark & \xmark & \xmark & \begin{tabular}[c]{@{}r@{}}0.272\\ (0.007)\end{tabular} & \begin{tabular}[c]{@{}r@{}}0.310\\ (0.002)\end{tabular} & \begin{tabular}[c]{@{}r@{}}1.195\\ (0.089)\end{tabular} & \begin{tabular}[c]{@{}r@{}}1.211\\ (0.005)\end{tabular} & \begin{tabular}[c]{@{}r@{}}1.716\\ (0.055)\end{tabular} & \begin{tabular}[c]{@{}r@{}}1.315\\ (0.018)\end{tabular} & \begin{tabular}[c]{@{}r@{}}0.918\\ (0.100)\end{tabular} & \begin{tabular}[c]{@{}r@{}}0.796\\ (0.031)\end{tabular} \\
\begin{tabular}[c]{@{}l@{}}MeRLOT\\ w.o. $\psi$\end{tabular} & \cmark & \cmark & \cmark & \xmark & 
\begin{tabular}[c]{@{}r@{}}-0.072\\(0.024)\end{tabular} & 
\begin{tabular}[c]{@{}r@{}}0.315\\(0.001)\end{tabular} & 
\begin{tabular}[c]{@{}r@{}}0.869\\(0.075)\end{tabular} & 
\begin{tabular}[c]{@{}r@{}}1.328\\(0.019)\end{tabular} & 
\begin{tabular}[c]{@{}r@{}}1.072\\(0.133)\end{tabular} & 
\begin{tabular}[c]{@{}r@{}}1.125\\(0.025)\end{tabular} & 
\begin{tabular}[c]{@{}r@{}}0.033\\(0.362)\end{tabular} & 
\begin{tabular}[c]{@{}r@{}}0.640\\(0.092)\end{tabular} \\
\bottomrule
\end{tabular}
\end{table*}

\begin{table*}[h]
\centering
\caption{Quantitative results on Omnipush. OOD task set combines the three out-of-distribution sets. The numbers for MLP and ANP are borrowed from~\protect\citet{bauza19omnipush}.
The experiment is repeated three times with different random seeds, and we reported the average of the runs with standard deviation.
}
\vspace{0.1in}
\begin{tabular}{llrrr}  
\toprule
Dataset & Methods & NLL & RMSE & Dist. Eq. \\
\midrule
\multirow{1}{*}{\begin{tabular}[c]{@{}l@{}}Omnipush\\ Test-set\end{tabular}}
 & MLP & 0.16 & .328 & 7.2 mm \\
 & ANP & -0.11 & \textbf{.225} & 4.9 mm \\
 & MetaFun & 1.59 (0.55) & .358 (.029) & 7.9 (0.6) mm \\
 & MeRLOT & \textbf{-4.17 (0.04)} & .232 (.002) & 5.1 (0.0) mm \\
\midrule
OOD 
 & MLP & 2.46 & .512 & 11.2 mm \\
 & ANP & 2.33 & .469 & 10.3 mm \\
 & MetaFun & 11.70 (2.89) & .495 (.019) & 10.9 (0.4) mm \\
 & MeRLOT  & \textbf{-3.21 (0.02)} & \textbf{.425 (.003)} & 9.3 (0.1) mm \\
 \midrule
\multirow{1}{*}{\begin{tabular}[c]{@{}l@{}}New\\ surface\end{tabular}}
 & MLP & 1.85 & .333 & 7.3 mm\\
 & ANP & 1.16 & .285 & 6.2 mm \\
 & MetaFun & 11.97 (0.52) & .361 (.023) & 7.9 (0.5) mm \\
 & MeRLOT  & \textbf{-4.14 (0.02)} & \textbf{.237 (.003)} & 6.0 (0.1) mm \\
 \midrule
\multirow{1}{*}{\begin{tabular}[c]{@{}l@{}}New\\ object\end{tabular}}
 & MLP & 2.80 & .601 & 13.2 mm\\
 & ANP & 3.09 & .558 & 12.2 mm\\
 & MetaFun & 22.87 (10.09) & .610 (.018) & 13.4 (0.4) mm \\
 & MeRLOT  & \textbf{-2.49 (0.17)} & \textbf{.535 (.007)} & 11.7 (0.2) mm \\
 \midrule
\multirow{1}{*}{Both} 
 & MLP & 2.72 & .562 & 12.3 mm\\
 & ANP & 2.73 & .517 & 11.3 mm\\
 & MetaFun & 11.56 (4.12) & .551 (.017) & 12.1 (0.4) mm \\
 & MeRLOT  & \textbf{-2.75 (0.14)} & \textbf{.486 (.004)} & 10.7 (0.1) mm \\
\bottomrule
\end{tabular}
\label{tab:omnipush_full}
\end{table*}

\end{document}